\documentclass[letterpaper]{article} 
\usepackage{aaai25}  
\usepackage{times}  
\usepackage{helvet}  
\usepackage{courier}  
\usepackage[hyphens]{url}  
\usepackage{graphicx} 
\usepackage{natbib}  
\usepackage{caption} 
\usepackage{amsfonts,amssymb}
\usepackage{amsmath}
\usepackage{mathtools}
\usepackage{booktabs}
\usepackage{subcaption}
\usepackage{cleveref}

\usepackage{algorithm}
\usepackage{algorithmic}

\usepackage{newfloat}
\usepackage{listings}
\DeclareCaptionStyle{ruled}{labelfont=normalfont,labelsep=colon,strut=off} 
\floatstyle{ruled}
\newfloat{listing}{tb}{lst}{}
\floatname{listing}{Listing}
\title{Defending Against Sophisticated Poisoning Attacks with \\ RL-based Aggregation in Federated Learning}
\author{
    Yujing Wang\textsuperscript{\rm 1,\rm 2},  
    Hainan Zhang\textsuperscript{\rm 1,\rm 2}\thanks{\;Corresponding author.}, 
    Sijia Wen\textsuperscript{\rm 1,\rm 2}, 
    Wangjie Qiu\textsuperscript{\rm 1,\rm 2},
    Binghui Guo\textsuperscript{\rm 1}
}
\affiliations{
    \textsuperscript{\rm 1}Beijing Advanced Innovation Center for Future Blockchain and Privacy Computing \\
    \textsuperscript{\rm 2} Institute of Artificial Intelligence, Beihang University, China\\
    \{wangyujing, zhanghainan\}@buaa.edu.cn
}

\usepackage{bibentry}

\begin{document}

\maketitle

\begin{abstract}
Federated learning is susceptible to model poisoning attacks, especially those meticulously crafted for servers.
Traditional defense methods mainly focus on updating assessments or robust aggregation against manually crafted myopic attacks. When facing advanced attacks, their defense stability is notably insufficient. Therefore, it is imperative to develop adaptive defenses against such advanced poisoning attacks.
We find that benign clients exhibit significantly higher data distribution stability than malicious clients in federated learning in both CV and NLP tasks. 
Therefore, the malicious clients can be recognized by observing the stability of their data distribution.
In this paper, we propose AdaAggRL, an RL-based Adaptive Aggregation method, to defend against sophisticated poisoning attacks. Specifically, we first utilize distribution learning to simulate the clients' data distributions. Then, we use maximum mean discrepancy (MMD) to calculate the pairwise similarity of the current local model data distribution, its historical data distribution, and global model data distribution. Finally, we use policy learning to adaptively determine the aggregation weights based on the above similarities.
Experiments on four real-world datasets demonstrate that the proposed defense model significantly outperforms widely adopted defense models for sophisticated attacks. 
\end{abstract}

%

\section{Introduction}
Federated Learning (FL) enables distributed model training across local devices, preserving data privacy while leveraging diverse local data to improve performance. It is widely applied in areas including smart healthcare~\cite{healthcare}, financial services~\cite{finance}, IoT~\cite{IoT}, and intelligent transportation~\cite{transportation}. However, FL systems are vulnerable, and the performance of the aggregated model is susceptible to model poisoning attacks from unknown clients~\cite{zheng2024safely}, especially the sophisticated poisoning strategies tailored for central servers. In this work, we focus on untargeted model poisoning attacks, where malicious devices aim to maximally reduce the accuracy of the global model by sending customized gradients to the server.

\begin{figure}[!t]
    \centering
    \begin{subfigure}{0.41\columnwidth}
        \includegraphics[width=1\columnwidth]{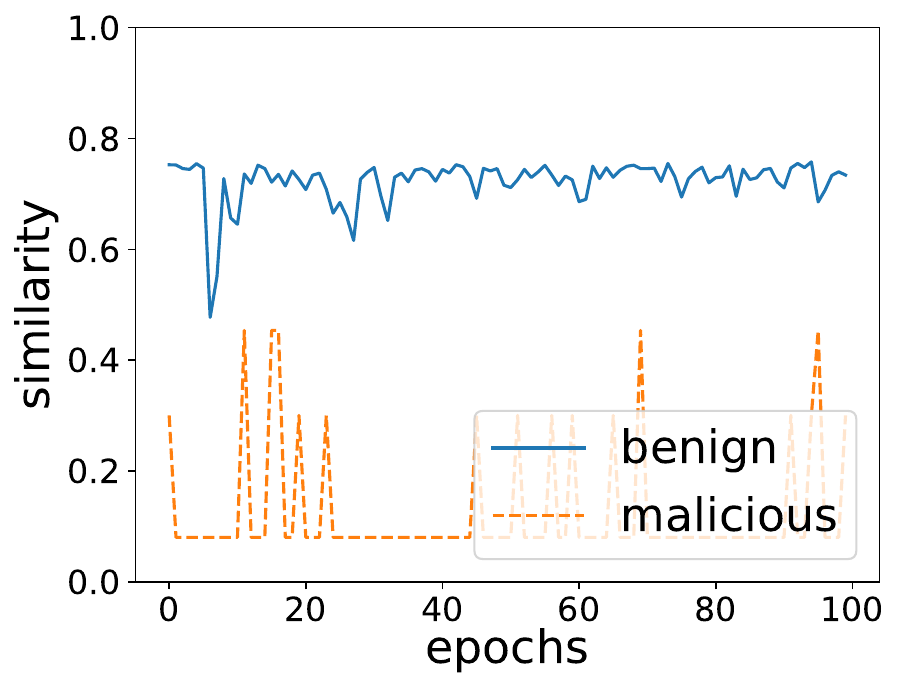}
        \caption{IPM}
    \end{subfigure}
    \begin{subfigure}{0.41\columnwidth}
        \includegraphics[width=1\columnwidth]{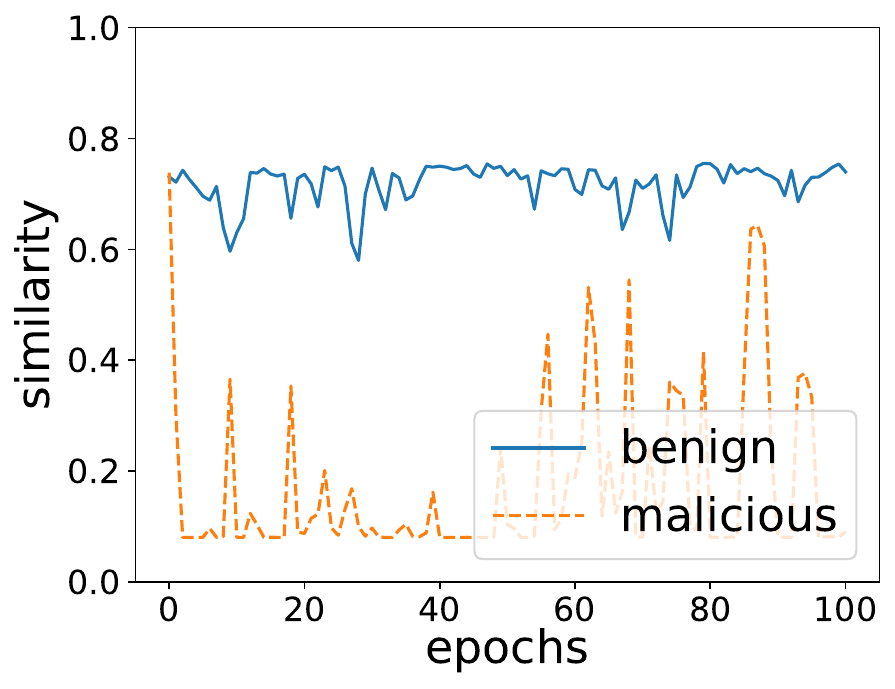}
        \caption{LMP}
    \end{subfigure}
    \begin{subfigure}{0.41\columnwidth}
    \includegraphics[width=1\columnwidth]{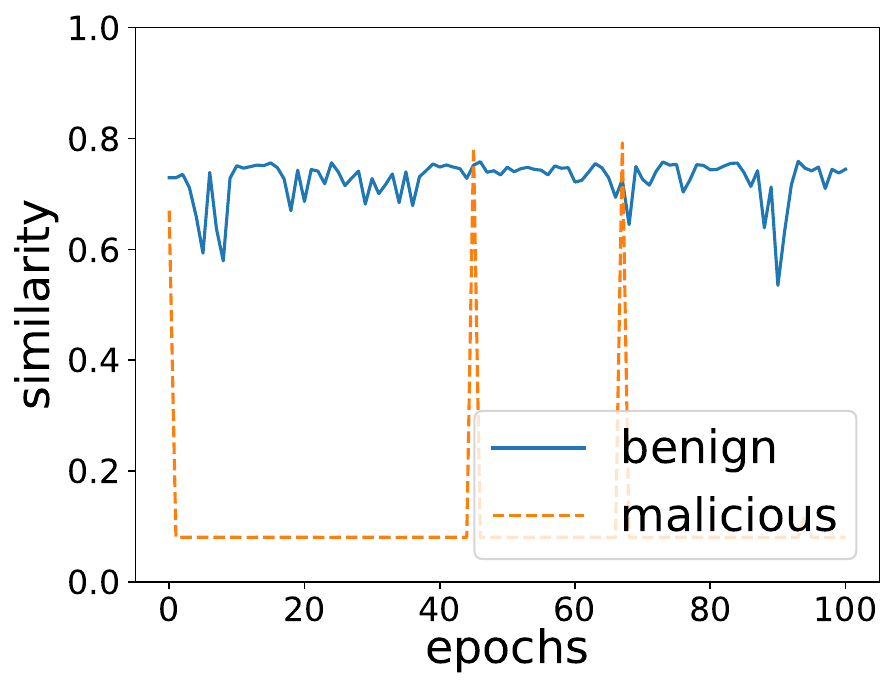}
    \caption{EB}
    \end{subfigure}
    \begin{subfigure}{0.41\columnwidth}
    \includegraphics[width=1\columnwidth]{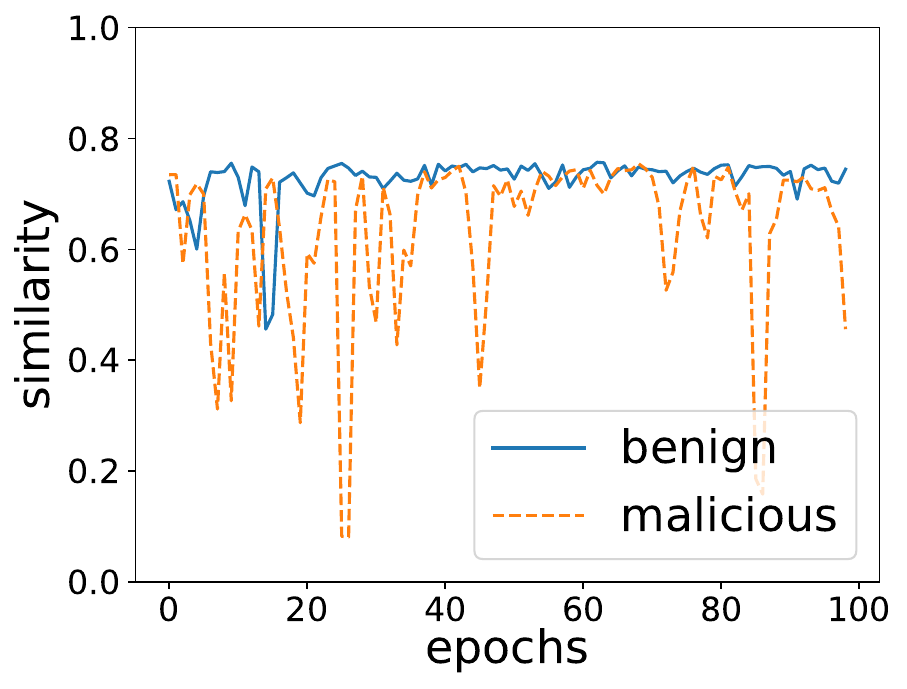}
    \caption{RL-attack}
    \end{subfigure}
    \caption{The statistical results of the similarity between the current client data distribution and its historical data distributions under four types of attacks vary with the training epochs on MNIST dataset.
    The x-axis denotes the number of client update rounds, and the y-axis represents the similarity between the current and its historical data distributions.}
    \label{fig:statistic}
\end{figure}

Traditional defense methods mainly rely on designing local model update assessment mechanisms or using robust aggregation methods to mitigate the impact of poisoning attacks. However, these defense methods are primarily targeted at manually crafted myopic attack strategies, and their defense stability is lacking when facing advanced attacks. For example, \citeauthor{RL-attack} propose using distribution learning to simulate the data distribution of the central server and employing reinforcement learning (RL) to tailor attack policy for the aggregation process, making it less detectable.
Therefore, it is urgent to develop adaptive defenses against such learnable advanced poisoning attacks.

Most benign clients exhibit significant data distribution stability in FL. In normal cases, the current training data distribution simulated by the client's parameters should align with its historical simulated data distribution. This is because benign clients typically employ random sampling and undergo multi-round training based on their local data, ensuring the stability of the simulated data distribution. However, malicious clients require gradient attacks, so its simulated data distribution between current and history lacks regularity. To validate this, we conduct a statistical analysis of mainstream attack methods, namely IPM~\cite{IPM}, LMP~\cite{LMP}, EB~\cite{EB} and RL-based attacks~\cite{RL-attack}, measuring the similarity of their data distribution with history after each round of updates on MNIST, as shown in Figure~\ref{fig:statistic}. We find that benign clients maintain higher data distribution similarity, while attack clients show no discernible patterns. We also observe the same phenomenon on NLP tasks, as shown in Appendix. Therefore, malicious clients can be recognized through the stability of their simulated data distribution.

This paper proposes an RL-based Adaptive Aggregation method, AdaAggRL, to thwart sophisticated poisoning attacks. It determines aggregation weights of local models by comparing the stability of client data distributions. Specifically, we first use distribution learning to emulate client data distributions based on the uploaded model parameters. Next, we use the maximum mean discrepancy (MMD) to calculate the pairwise similarity of the current local model data distribution, its historical data distribution, and the global model data distribution, to evaluate the stability of the client's data distribution. Considering that the accuracy of distribution learning can potentially impact the calculation of the similarities above, we use reconstruction similarity as an evaluation metric for the quality of distribution learning. Finally, we use the policy learning method TD3 to adaptively determine the aggregation weights based on these similarities.

Experimental results on four real-world datasets demonstrate that AdaAggRL defense method significantly outperforms existing defense approaches~\cite{Krum,Median,clipping} and achieves a more stable global model accuracy even facing sophisticated attacks, such as RL-based attacks~\cite{RL-attack}.

The innovations of this paper are as follows:
\begin{itemize}
    \item We propose an RL-based adaptive aggregation method AdaAggRL to defend against sophisticated untargeted model poisoning attacks tailored for servers in FL, aiming to advance the development of defense systems.
    \item We observe stable training data distribution in benign clients over time, contrasting with irregular distribution caused by disruptive attempts from malicious clients in both CV and NLP tasks. Thus, we propose four metrics as RL environmental cues, utilizing policy learning to determine local model aggregation weights based on observed cue changes.
    \item  Experimental results on four datasets demonstrate that the proposed AdaAggRL defense method can maintain more stable global model accuracy than baselines, even when more advanced customized attacks are applied.
\end{itemize}

\section{Related Work}
\subsection{Poisoning Attacks}
Based on the attacker's objectives, poisoning attacks can be classified into targeted poisoning attacks aiming to misclassify specific input sets \cite{EB,NEURIPS2019_ec1c5914,bagdasaryan2020backdoor} and untargeted attacks aimed at reducing the overall accuracy of the global model \cite{LMP,IPM,shejwalkar2021manipulating}. Current untargeted attack methods typically employ heuristic-based approaches \cite{IPM} or optimize myopic objectives \cite{LMP,shejwalkar2021manipulating,shejwalkar2022back}. 
\citeauthor{EB} generate malicious updates through explicit enhancement, optimizing for a malicious objective strategically designed to induce targeted misclassification. \citeauthor{IPM} manipulate the attacker's gradients to ensure the inner product with the true gradients becomes negative. \citeauthor{LMP} generate malicious updates by solving an optimization problem. However, these attack methods require local updates from benign clients or accurate global model parameters to generate significant adversarial impacts and often yield suboptimal results when robust aggregation rules are employed.

To address these deficiencies, \citeauthor{RL-attack} propose a model-based RL framework for guiding untargeted poisoning attacks in FL system. They utilize the server's updates to approximate the server's data distribution, subsequently employing the learned distribution as a simulator for FL environment. Based on the simulator, they utilize RL to automatically generate effective attacks, resulting in significantly reducing the global accuracy. Even when the server employs robust aggregation rules, this customized method can still maintain a high level of attack effectiveness.

\subsection{Defenses for Poisoning Attacks}

Current defense strategies against FL poisoning attacks can be categorized into two types: one involves designing local model update evaluation mechanisms to identify malicious client-submitted model parameters~\cite{FLtrust, sattler2020byzantine,fldetector}, and the other is based on designing novel Byzantine fault-tolerant aggregation algorithms using mathematical statistics to improve the robustness of aggregation~\cite{Krum,Median,clipping,rajput2019detox,xie2019zeno}. 
In evaluation mechanisms, \citeauthor{sattler2020byzantine} divide updates into different groups based on cosine similarity between the model parameters submitted by clients, mitigating the impact of poisoning attacks. In response to more covert poisoning attacks, some evaluation methods~\cite{FLtrust,park2021sageflow} require the server to collect a portion of clean data as a basis for validating model updates. 
In robust aggregation algorithms, statistical methods~\cite{Krum,Median,clipping} compare local updates and remove statistical outliers before updating the global model. For example,~\citeauthor{Krum} employ the square-distance metric to measure distances among local updates and select the local update with the minimum distance as the global parameters. \citeauthor{Median} sort the values of parameters in all local model updates and consider the median value of each parameter as the corresponding parameter value in the global model update. \citeauthor{clipping} perform gradient clipping on parameter updates before aggregation. Due to the susceptibility of statistical estimates to an outlier, existing aggregation methods cannot guarantee accuracy well and are still susceptible to local model poisoning attacks \cite{EB,LMP}.

These defense methods mainly rely on the local model parameters. For sophisticated attacks, such as RL-based customized attacks, it is difficult to identify malicious updates from the parameter information alone accurately. Moreover, statistics-based robust aggregation is not flexible enough.

\section{Motivation}
During FL, benign clients are selected by the server through random sampling and trained on their local data for a certain number of rounds. Therefore, the simulated data distribution obtained through gradient inversion should align with their historical distribution. But the data distribution of malicious clients lacks regularity, due to their need to conduct model attacks. To verify this, we conduct a statistical analysis of the similarity between data distributions for mainstream attack methods such as IPM, LMP, EB, and RL-attack, as shown in Figure~\ref{fig:statistic}. The similarity of data distributions for benign clients remains high and stable, while malicious clients lack any regular pattern. 
We also observe the same phenomenon on NLP tasks, as shown in Appendix.

\begin{figure}[!t]
    \centering
    \begin{subfigure}{0.41\columnwidth}
        \includegraphics[width=1\columnwidth]{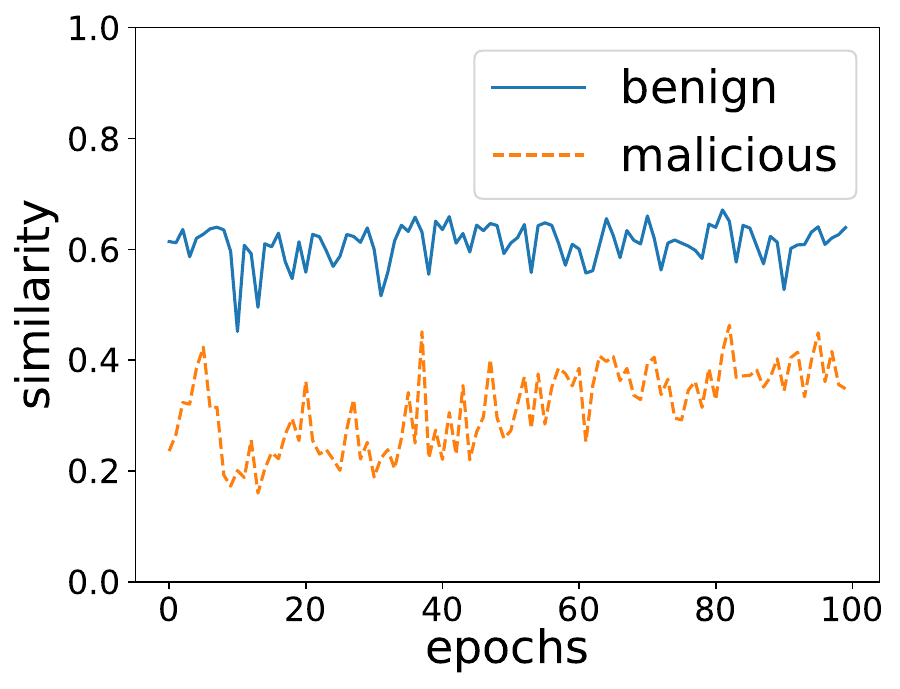}
        \caption{IPM}
    \end{subfigure}
    \begin{subfigure}{0.41\columnwidth}
        \includegraphics[width=1\columnwidth]{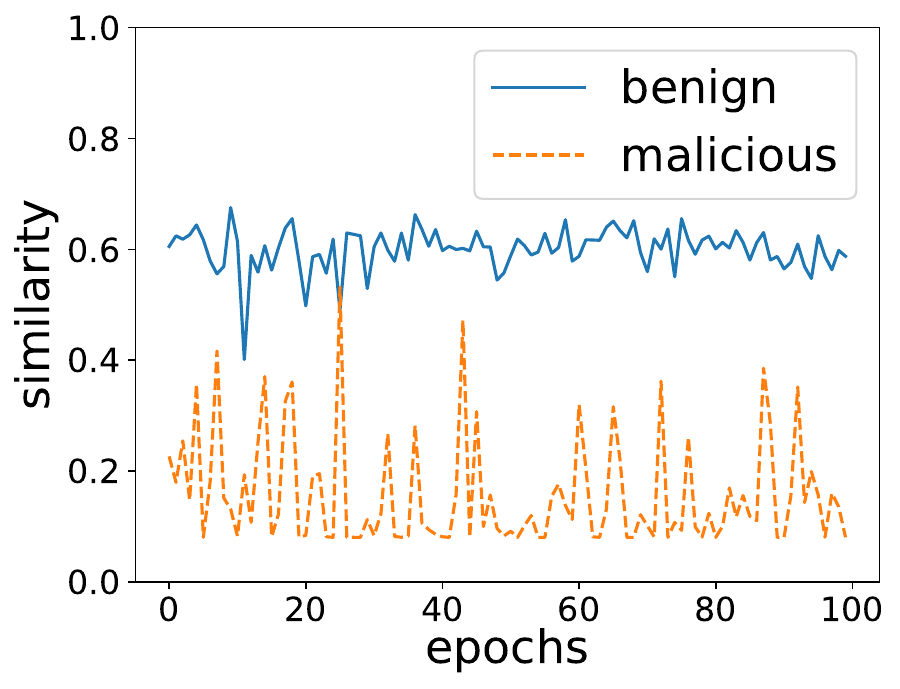}
        \caption{LMP}
    \end{subfigure}
    \begin{subfigure}{0.41\columnwidth}
        \includegraphics[width=1\columnwidth]{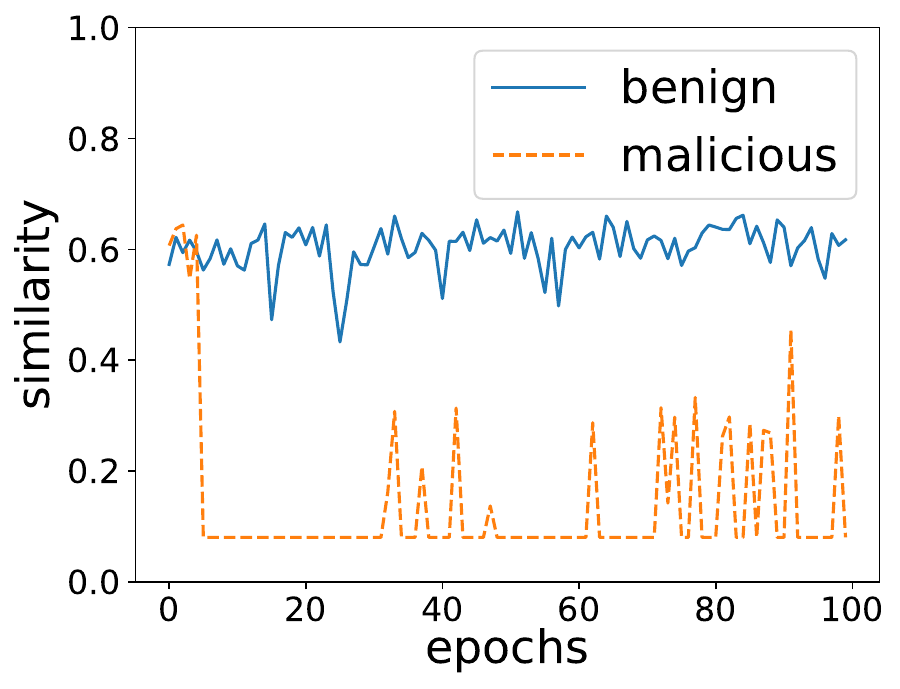}
        \caption{EB}
    \end{subfigure}
    \begin{subfigure}{0.41\columnwidth}
        \includegraphics[width=1\columnwidth]{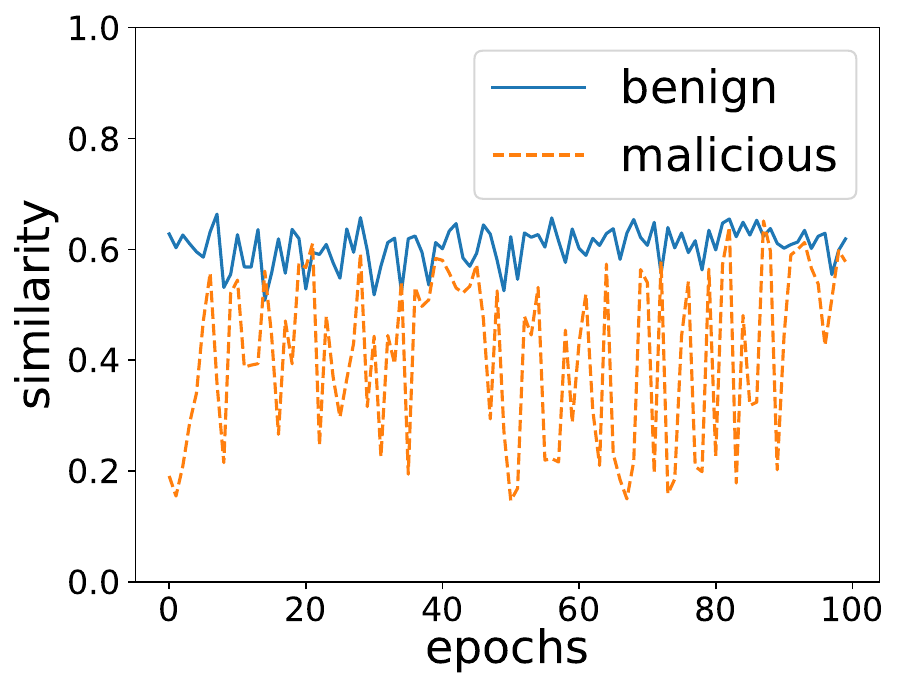}
        \caption{RL-attack}
    \end{subfigure}
    
    \caption{The statistical results of the similarity between the current client data distribution and the global model data distribution under attacks vary with epochs on MNIST.}

    \label{fig:2}
\end{figure}

Moreover, 
the data distribution of benign clients for current steps is similar to the global model and remains stable, but the similarity of malicious clients is lower, as shown in Figure~\ref{fig:2}. Since the data distribution of the global model represents the average state of normal data, the distribution of benign clients should always be consistent with the global model. Instead, malicious clients do not possess this property. 
Therefore, we can compare data distribution similarity between the current client and global model to observe variations and assess the degree of malicious behavior. Similarly, the similarity between historical distribution and global model distribution is higher and more stable for benign clients than malicious ones, as shown in Appendix. 

Since the quality of distribution learning greatly affects the accuracy of distribution similarities, the reconstruction similarity of distribution learning is used as an evaluation metric for assessing the quality of distribution learning. 
Reconstruction similarity measures the similarity between the inversion gradients from distribution learning and the gradients updated by clients. Higher reconstruction similarity indicates higher confidence in current distribution learning.

\section{RL-based Adaptive Aggregation Methods}
\subsection{Task Definition}
In the context of FL \cite{FL}, a system comprises $K$ clients, with each client $k$ possessing a fixed local dataset $D_k=\{(x_{kj}, y_{kj})\}_{j=1}^{n_k}$, where $n_k$ is the size of $D_k$. The local objective of client $k$ is $F_k(\theta) = \frac{1}{n_k} \sum_{j=1}^{n_k} l(\theta, x_{kj}, y_{kj})$, where $l$ is the loss function. And $(x_{kj}, y_{kj})$ is the $j$-th sample drawn i.i.d. from some distribution $P_k$. $\hat{P}_k$ denotes the empirical distribution of $n_k$ data samples. The optimization objective of FL is: $\min_{\theta} f(\theta)=\sum_{k=1}^{K} p_k F_k(\theta)$, $p_k$ represents the weight of client $k$. 
During FL, in each epoch $t \geq 0$, the server randomly selects a subset $C^t$ from all clients and distributes the latest global model parameters \(\theta^t\). The chosen clients train on their local datasets, updating parameters as \(\theta^{t+1}_k = \theta^t - \alpha \nabla F_k(\theta)\), where \(\alpha\) is the learning rate. Then they upload the updated model parameters. The server aggregates the received parameters according to a specific aggregation rule \(Aggr\), to obtain the new global model parameters \(\theta^{t+1} = Aggr(\theta^{t+1}_{C^t})\).

We assume that the server is non-malicious. 
Malicious clients have sufficient knowledge of the server, including model structure, loss function, learning rate, and other key parameters, to demonstrate the effectiveness of AdaAggRL in defending against strong attacks.

\subsection{Framework Overview}
    
\begin{figure}
    \centering
    \includegraphics[width=0.90\linewidth]{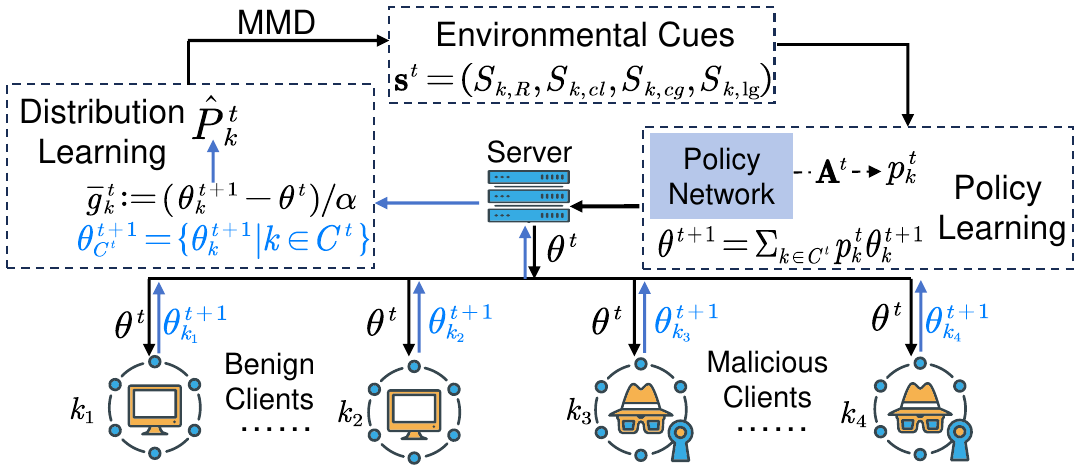}
    \caption{An overview of our AdaAggRL}
    \label{fig:overview}
\end{figure}
In AdaAggRL framework (see Algorithm in Appendix), the server determines the weights for local model aggregation by assessing the stability of client data distributions, as shown in Figure \ref{fig:overview}. Firstly, we employ distribution learning to simulate the client's data distribution $\hat{P}^t_k$ based on the locally uploaded model parameters $\theta^{t+1}_k$. Secondly, we calculate similarity metrics $S_{k,cl}$ between the current client and historical data distribution, $S_{k,cg}$ between the current client and global model data distribution, and $S_{k,lg}$ between the client's historical and global model data distribution. Finally, RL is utilized to adaptively determine the weight $p_k$ for local model aggregation based on these metrics and the reconstruction similarity $S_{k,R}$ from distribution learning.
\subsection{Distribution Learning}
According to local model parameters $\theta^{t+1}_k$ uploaded by clients, the server simulates the local data distribution $\hat{P}^t_k$ of clients by gradient inversion. In this work, we adapt the inverting gradients (IG) method~\cite{invertgradient} for distribution learning. The IG method reconstructs the data sample by optimizing the loss function based on the cosine similarity between the real gradient and the gradient generated by the reconstructed data. 

For each epoch $t\geq 0$, the server receives the clients’ locally updated model parameters and calculates the corresponding batch level gradient $\bar{g}^t_k\coloneqq(\theta^{t+1}_k-\theta^{t})/\alpha$. The server then solves the following optimization problem with a batch of randomly generated dummy data and labels $D_{dummy}$: $\min_{D_{\text{dummy}}} 1 - \frac{\langle \nabla_{\theta}F_{D_{\text{dummy}}}(\theta^{t+1}_k), \bar{g}^t_k \rangle}{\|\nabla_{\theta}F_{D_{\text{dummy}}}(\theta^{t+1}_k)\| \cdot \|\bar{g}^t_k\|}
+\frac{\beta}{B'}\Sigma_{(x,y)\in D_{\text{dummy}}}\mathrm{TV}(x)$, where $\beta$ is a fixed parameter, $B'$ is the size of the dummy data batch, $F_{D_{\text{dummy}}}(\theta)=\frac{1}{B'} \sum_{(x,y)\in D_{\text{dummy}}} l(\theta, x, y)$, $\mathrm{TV}$ calculates the total variation\cite{rudin1992nonlinear}. While solving the optimization problem, $D_{dummy}$ is continuously updated. 
The optimization terminates after $max\_iters$ iterations, then outputs the updated data as the reconstructed data samples $D_{\text{rec}}$, and the reconstruction similarity of client $k$, denoted as $S_{k,R}=\frac{\langle \nabla_{\theta}F_{D_{\text{rec}}}(\theta^{t+1}_k), \bar{g}^t_k \rangle}{\|\nabla_{\theta}F_{D_{\text{rec}}}(\theta^{t+1}_k)\| \cdot \|\bar{g}^t_k\|}$.
\subsection{Environmental Cues}

The distribution of samples is extracted by employing a pre-trained CNN to convert the image samples $D_{\text{rec}}$ into a collection of feature vectors $V$. For each client $k$, the difference $d_{k,cl}$ between this feature distribution and the history distribution is calculated using the maximum mean discrepancy (MMD) \cite{arbel2019maximum,wang2021rethinking} as $d_{k,cl}=\mathrm{MMD}(V^{\text{current}}_{k},V^{\text{history}}_{k})\in[0,+\infty)$. 
Here, $V^{\text{current}}_k$ and $V^{\text{history}}_k$ are feature vectors of current and historical data distributions respectively. $V_g$ represents feature vectors of the global model data distribution, obtained by averaging feature vectors from all participating clients. 
Subsequently, the dissimilarity between the current data distribution of client $k$ and the global one is $d_{k,cg}=\mathrm{MMD}(V^{\text{current}}_{k},V_g)$. Similarly, the dissimilarity between the historical data distribution and the global one is $d_{k,lg}=\mathrm{MMD}(V^{\text{history}}_{k},V_g)$. And the similarity between the current data distribution and the historical data distribution $S_{k,cl}$ is calculated using the following formula: 
\begin{equation}
    S_{k,cl}=
    2\cdot\cos(\tanh(\frac{d_{k,cl}}{2}))-1
\end{equation}
So $S_{k,cl}\in(0,1]$, and as the current data distribution obtained through gradient inversion becomes more consistent with the historical data distribution, the value of $S_{k,cl}$ increases. Therefore, we can determine $S_{k,cg}$ between the current client data distribution and the global model data distribution, as well as $S_{k,lg}$ between the historical client data distribution and the global model data distribution.
\subsection{Actions Learning}

The server dynamically adapts the aggregation weights based on three similarity metrics and the reconstruction similarity associated with each client. By utilizing experiences sampled from the simulated environment, the server engages in learning a collaborative defense strategy aimed at minimizing empirical loss. This learning process employs the RL algorithm TD3 \cite{TD3}.

\textbf{State:} To simulate the training process of FL, including malicious clients and their behaviors, an environment for RL is set up. For each epoch \(t\) in FL, let \(\mathbf{s}^t = (s^t_{k_1}, s^t_{k_2}, ..., s^t_{k_{|C^t|}})^T\in (0,1]^{|C^t|\times 4}\) be the state of the reinforcement learning simulation environment, where \(k_j\) denotes the client identifier participating in the training. Here, \(s^t_k := (S_{k,R}, S_{k,cl}, S_{k,cg}, S_{k,lg})\) represents the state of client \(k\) with the reconstruction similarity and three obtained metrics. So the state search space is $(0,1]^{|C^t|\times 4}$, where $|C^t|$ is the number of clients participating in federated aggregation.

\textbf{Action:} Through RL, the server is trained to make the decision \(\mathbf{A}^t = (\mathbf{a}^t, b^t)^T\in [0,1]^5\) based on the current FL environment state \(\mathbf{s}^t\). Here, \(\mathbf{a}^t\) is a four-dimensional vector, and \(\Sigma_{i=1}^4 a^t_i = 1\), where \(a^t_i \in [0,1]\) represents the weighted weight for four environmental parameters $(S_{k,R}, S_{k,cl}, S_{k,cg}, S_{k,lg})$, and \(b^t \in [0,1]\) represents a threshold. So the action space is $[0,1]^5$.

\textbf{State transition:} The FL system changes based on the server's decision \(\mathbf{A}^t\). Specifically, the FL system first obtains \(\hat{\mathbf{w}} = \mathbf{s}^t \cdot \mathbf{a}^t \in \mathbb{R}^{|C^t|}\), i.e., weighting the environmental parameters. $\hat{w}_k=s^t_k\cdot\mathbf{a}^t$ indicates the score of client $k$. Then, function \(g(\cdot)\) maps \(\hat{\mathbf{w}}\) to $[0,1]$ interval and normalizes it, resulting in \(\tilde{\mathbf{w}} = g(\hat{\mathbf{w}})\). Let \(\delta = \max(\tilde{\mathbf{w}}) \cdot b^t\), define \begin{equation}
    f_{\delta}(x) = 
\begin{cases} 
      x & \text{if } x > \delta \\
      0 & \text{if } x \leq \delta 
\end{cases}
\end{equation}
Then \(\mathbf{w} = f_{\delta}(\tilde{\mathbf{w}})\). The function \(f_\delta\) denotes clients with lower scores, which are considered to exhibit malicious behavior and are excluded from aggregation. The generation process of the server policy reveals that the threshold \(\delta\) within \(f_\delta\) is also adaptively adjusted based on the state. 

Additionally, a vector \(\mathbf{h}^t\in\mathbb{N}^K\) is introduced in the FL system environment to record the malicious behaviors of each client. $h^0_k=0$, when \(\tilde{w}_k \leq \delta\), \(h^{t+1}_k = h^t_k + 1\), otherwise, \(h^{t+1}_k = \max(h^t_k - 1,0)\), representing the occurrence of malicious behavior by client $k$. Based on these outcomes, the FL system determines the aggregation strategy for global model parameters of the new round as follows, \begin{equation}
    \theta^{t+1}=\Sigma_{k\in C^t}\frac{w_k}{\lambda^{h^{t+1}_k}}\cdot\theta^{t+1}_k
\end{equation}
$\lambda\in[1,+\infty)$ is a hyperparameter used to indicate the severity of the penalty for malicious behavior. A higher value of \(\lambda\) corresponds to a stronger punitive impact. In the subsequent epoch \(t+1\), the FL system selects a new subset of clients for training, denoted as \(C^{t+1}\), and disseminates the updated model parameters \(\theta^{t+1}\) to the clients. Each client then locally trains on its dataset to obtain local parameters \(\theta^{t+1}_k\), resulting in a new state \(\mathbf{s}^{t+1}\).

\textbf{Reward:} The FL system calculates the reward at step \(t\) as \(r\coloneqq f(\theta^t) - f(\theta^{t+1})\) based on the newly obtained global model parameters \(\theta^{t+1}\). 

\begin{table}[tbp]
\begin{center}
\begin{footnotesize}
\setlength{\tabcolsep}{1mm}
\begin{tabular}{lccccccc}
\toprule
    & Our & Krum & Median & C-Median & FLtrust & Clipping \\
\midrule
MNIST            & 1.652& 1.744& 1.069& 1.556& 1.107 & 1.327 \\
Cifar10          & 7.201& 8.698& 3.537& 4.469& 3.589& 3.696\\
\bottomrule
\end{tabular}
\end{footnotesize}
\end{center}
\caption{The training time of different algorithms for one round of FL (s)}
\label{table:timecost}
\end{table}
\subsection{Computational Complexity}
Compared to classical FL, AdaAggRL's increased computational complexity mainly stems from $\mathrm{EnvironmentalCues}$ step. In IG algorithm, computation involves model forward propagation, loss function calculation, and backpropagation for optimization. The computational complexity depends on model complexity $M$, optimization rounds $max\_iters$, and reconstructed images $num\_images$, totaling $O(max\_iters \times M \times num\_images)$. Extracting image features via pre-trained CNN incurs a complexity of $O(C_{CNN} \times num\_images)$. Considering MMD's constant complexity $O(C_{MMD})$, the overall computational complexity for $\mathrm{EnvironmentalCues}$ is $O(|C^t| \times num\_images \cdot (M \times max\_iters + C_{CNN})) + O(3|C^t| \cdot C_{MMD}) + O(1)$.

Gradient inversion simulates data distributions, but the server's goal isn't precise image reconstruction. Thus, in our experiments, gradient inversion optimization is restricted to 30 steps, with 16 dummy images. Table~\ref{table:timecost} compares AdaAggRL's training time per FL round with other algorithms. AdaAggRL's time increases by an average of 21.4\% on MNIST and 50.1\% on CIFAR-10 compared to baselines.
\section{Experiments}

\begin{figure}[ht]
    \centering
    \begin{subfigure}{0.43\linewidth}
        \includegraphics[width=1\linewidth]{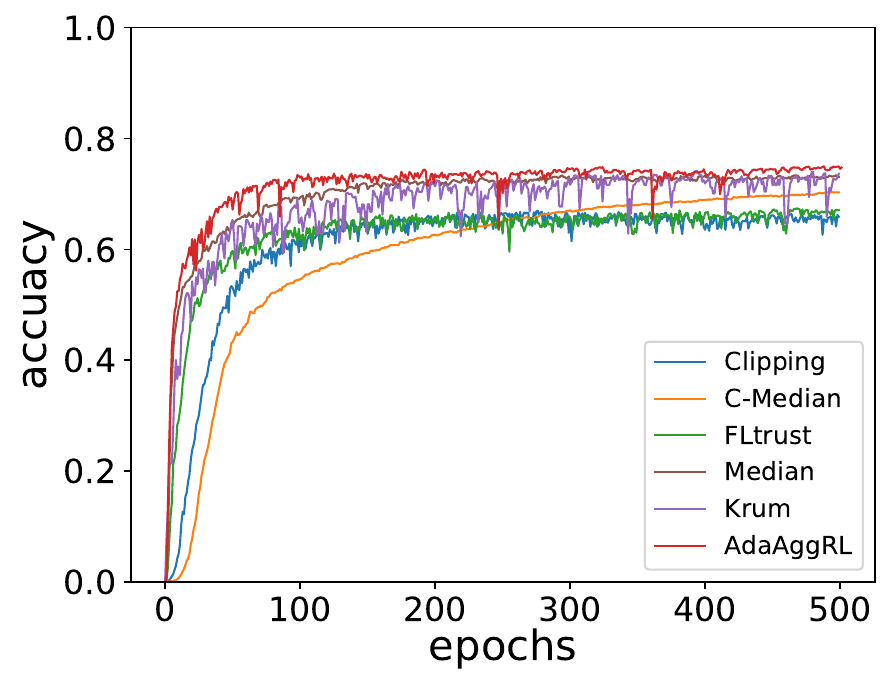}
        \caption{IPM}
    \end{subfigure}
    \begin{subfigure}{0.43\linewidth}
        \includegraphics[width=1\linewidth]{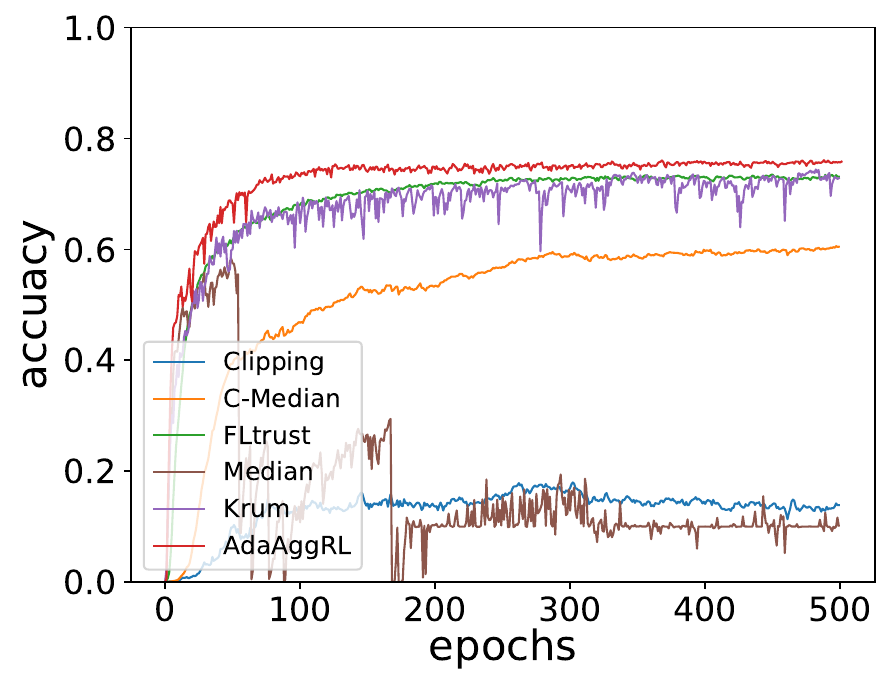}
        \caption{LMP}
    \end{subfigure}
    \begin{subfigure}{0.43\linewidth}
        \includegraphics[width=1\linewidth]{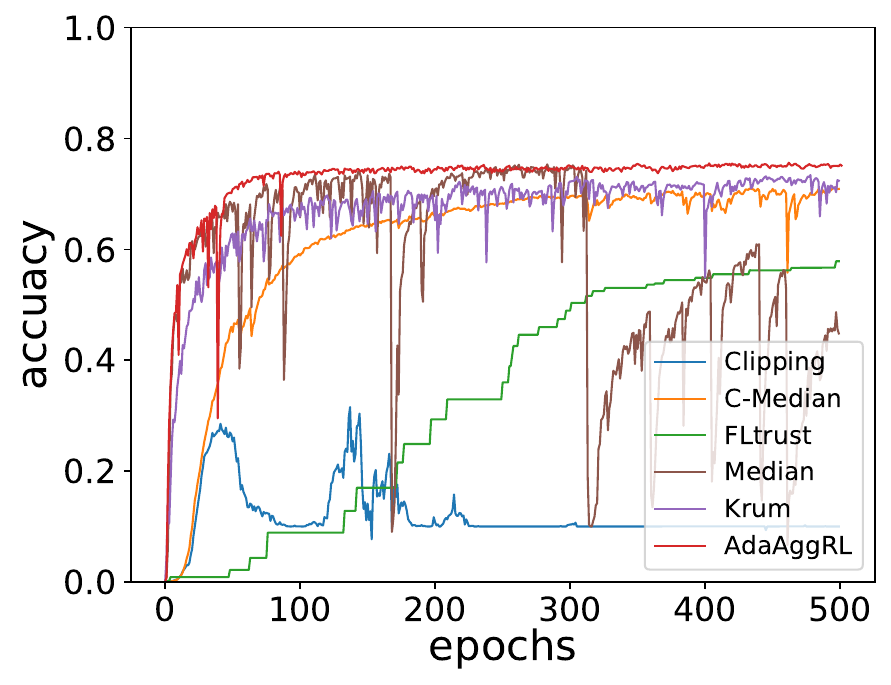}
        \caption{EB}
    \end{subfigure}
    \begin{subfigure}{0.43\linewidth}
        \includegraphics[width=1\linewidth]{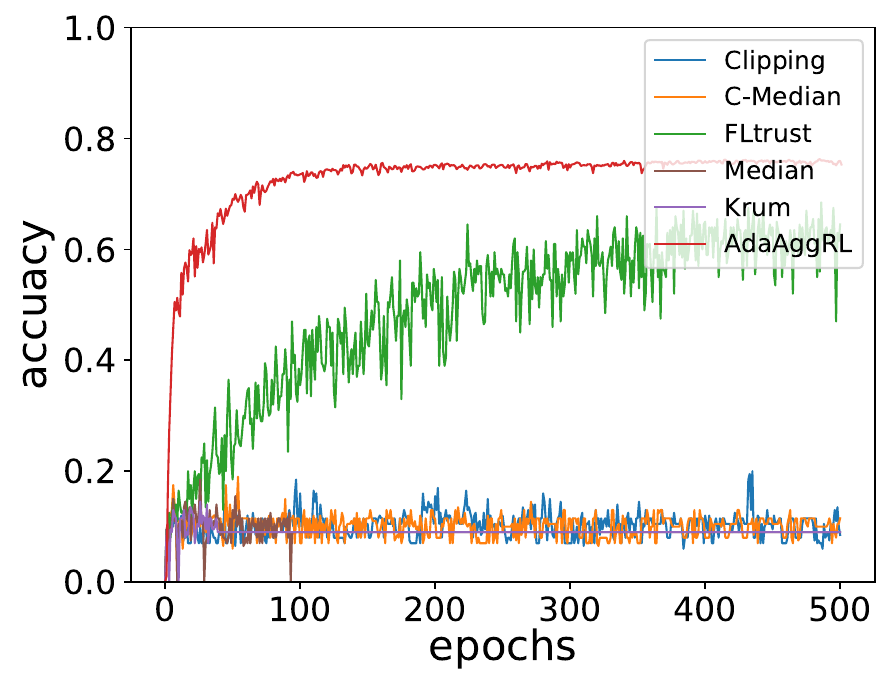}
        \caption{RL-attack}
    \end{subfigure}
    \caption{The testing accuracy variation of the global model on Cifar10 dataset under four attacks.}
    \label{fig:cifar10}
    
\end{figure}
\begin{figure}[t]
    \centering
    \begin{subfigure}{0.41\linewidth}
        \includegraphics[width=1\linewidth]{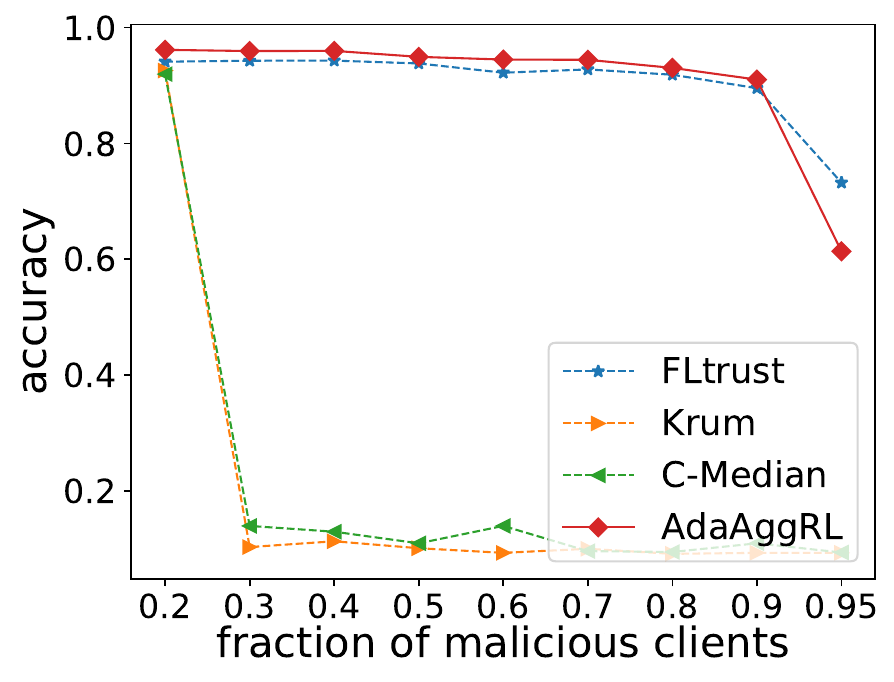}
        \caption{LMP}
        \label{attacker_lmp}
    \end{subfigure}
    \begin{subfigure}{0.41\linewidth}
        \includegraphics[width=1\linewidth]{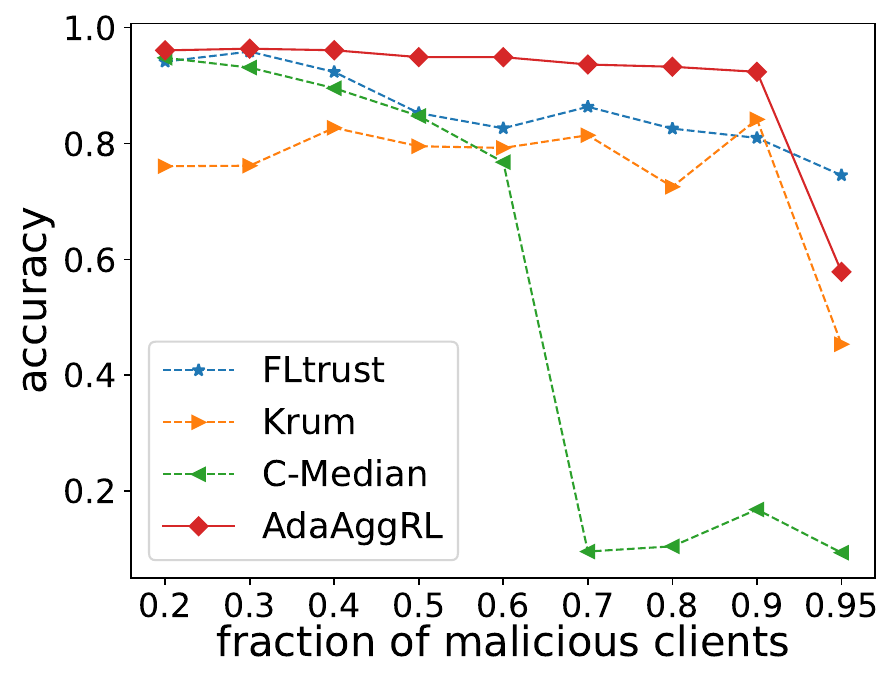}
        \caption{EB}
        \label{attacker_eb}
        
    \end{subfigure}
   \begin{subfigure}{0.41\linewidth}
        \includegraphics[width=1\linewidth]{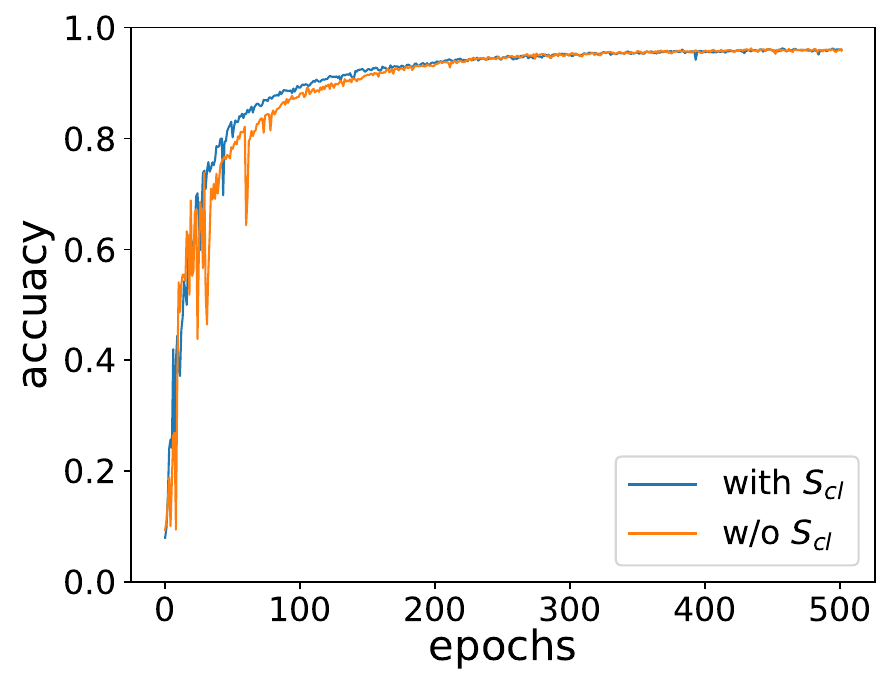}
        \caption{LMP}
        \label{Scl}
    \end{subfigure}
    \begin{subfigure}{0.41\linewidth}
        \includegraphics[width=1\linewidth]{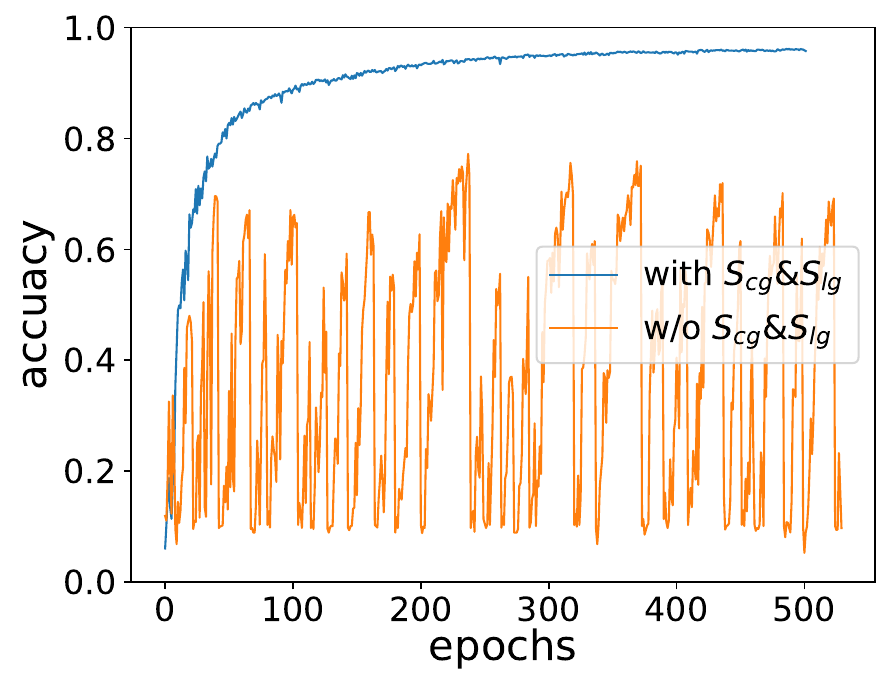}
        \caption{EB}
        \label{Scg}
    \end{subfigure}
    \caption{The testing accuracy of FL methods on MNIST-0.5 under LMP and EB as the proportion of malicious clients increases (a-b). The defense performance of AdaAggRL on MNIST-0.5 compared to the case where $S_{cl}$ is not considered under LMP (c) and the case where $S_{cg}$ and $S_{lg}$ are not considered under EB (d).}
    \label{fig:ablation}   
\end{figure}
\begin{figure*}[t]
    \centering
    \begin{subfigure}{0.22\linewidth}
        \includegraphics[width=1\linewidth]{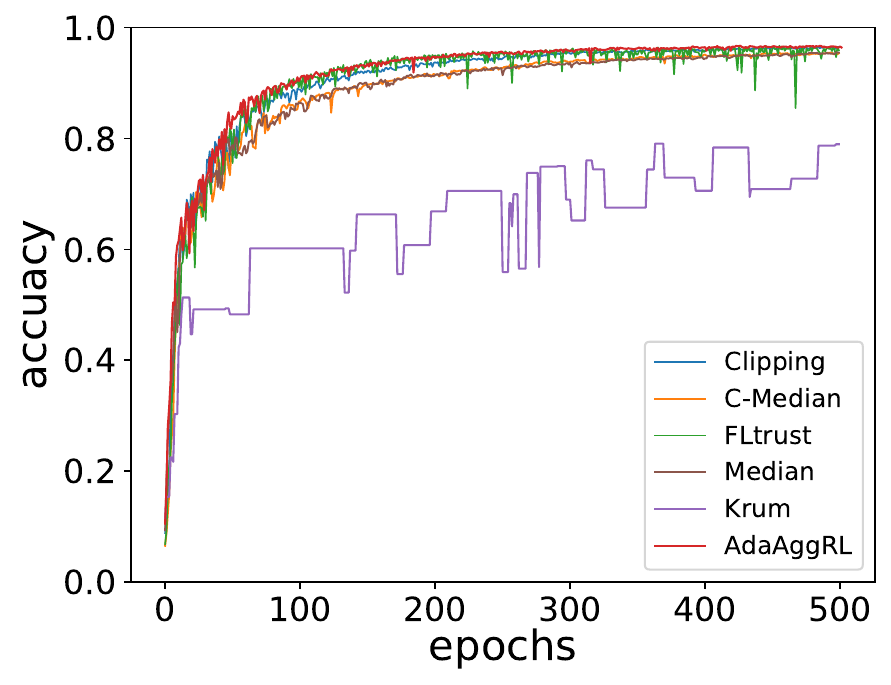}
        \caption{MNIST-0.1, IPM}
    \end{subfigure}
    \begin{subfigure}{0.22\linewidth}
        \includegraphics[width=1\linewidth]{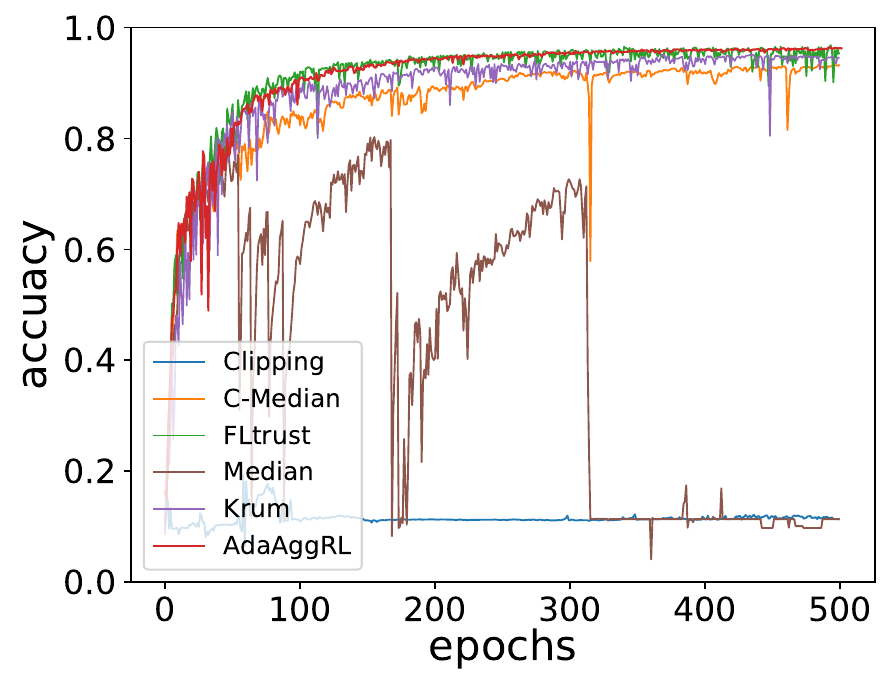}
        \caption{MNIST-0.1, LMP}
    \end{subfigure}
    \begin{subfigure}{0.22\linewidth}
        \includegraphics[width=1\linewidth]{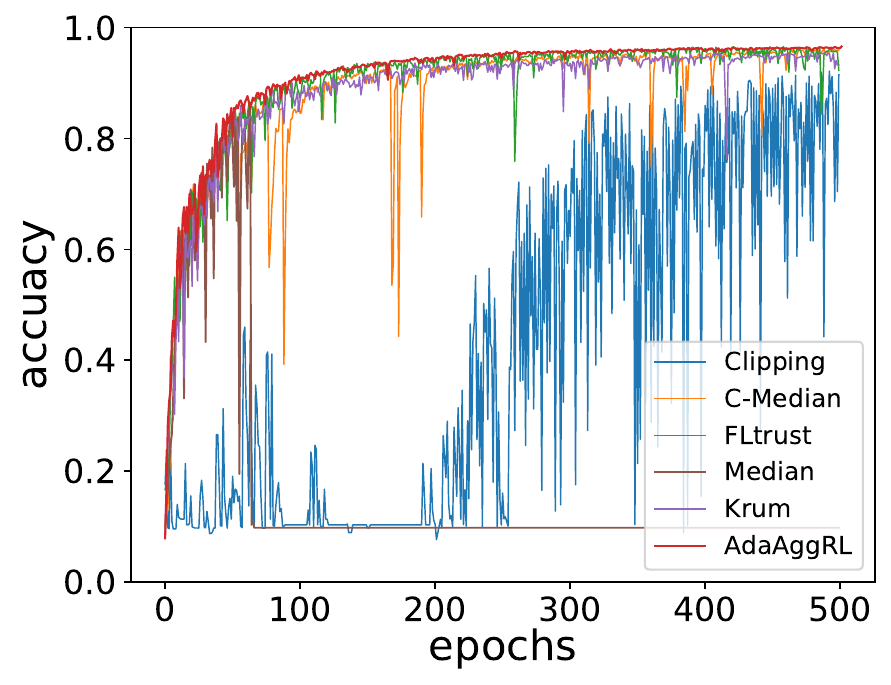}
        \caption{MNIST-0.1, EB}
    \end{subfigure}
    \begin{subfigure}{0.22\linewidth}
        \includegraphics[width=1\linewidth]{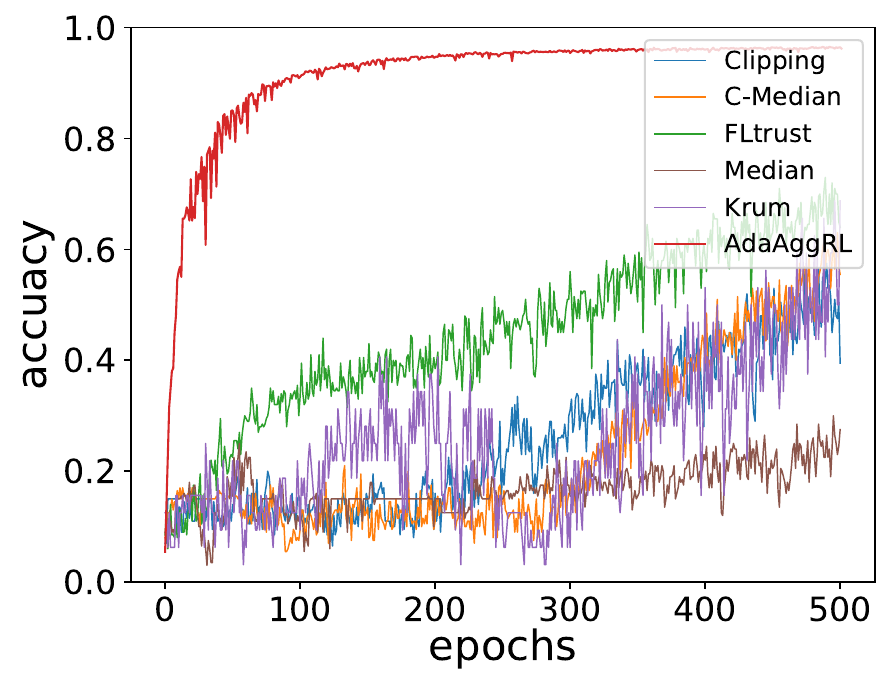}
        \caption{MNIST-0.1, RL-attack}
    \end{subfigure}

    \begin{subfigure}{0.22\linewidth}
        \includegraphics[width=1\linewidth]{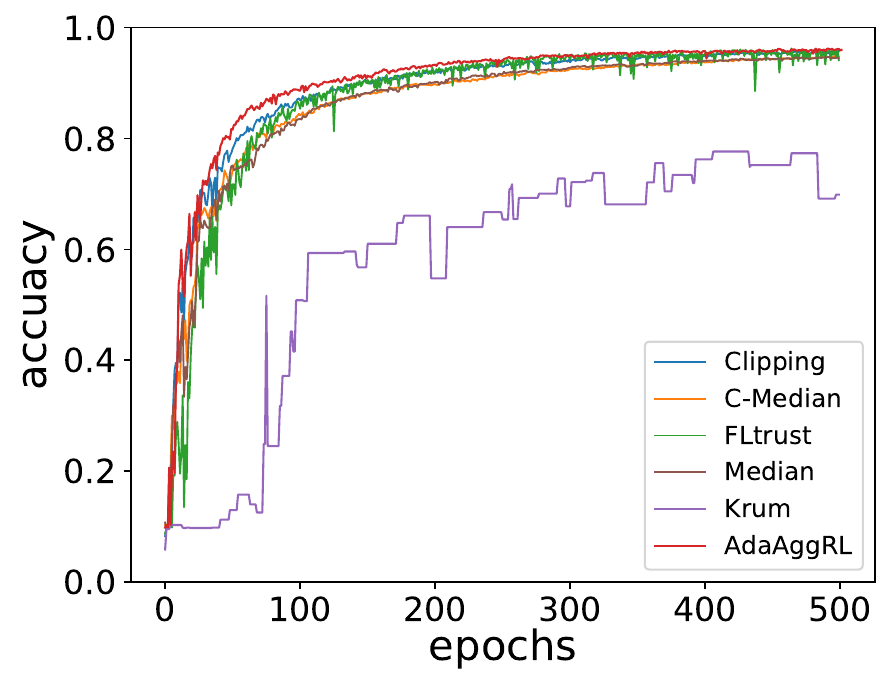}
        \caption{MNIST-0.5, IPM}
    \end{subfigure}
    \begin{subfigure}{0.22\linewidth}
        \includegraphics[width=1\linewidth]{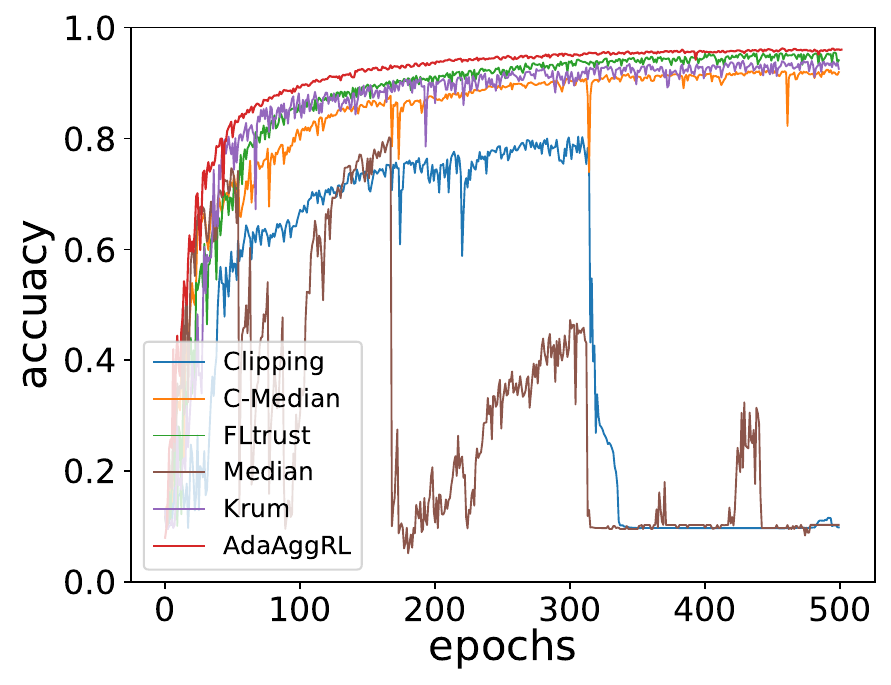}
        \caption{MNIST-0.5, LMP}
    \end{subfigure}
    \begin{subfigure}{0.22\linewidth}
        \includegraphics[width=1\linewidth]{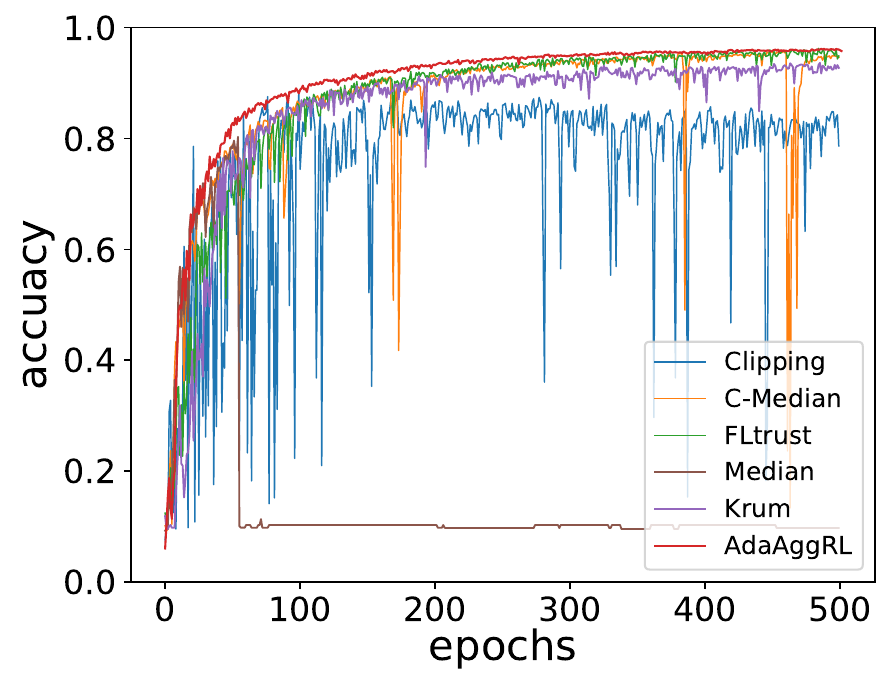}
        \caption{MNIST-0.5, EB}
    \end{subfigure}
    \begin{subfigure}{0.22\linewidth}
        \includegraphics[width=1\linewidth]{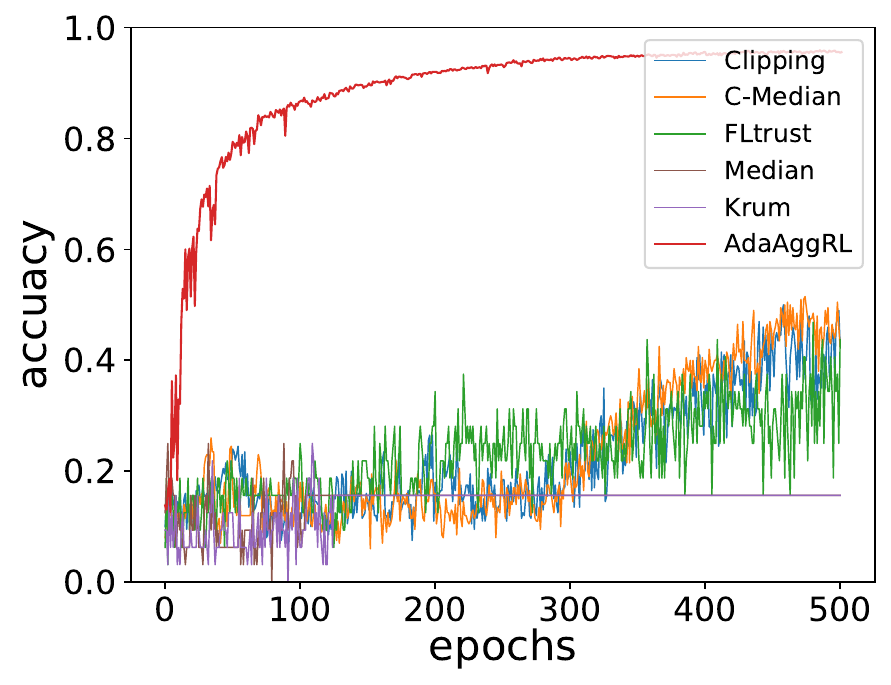}
        \caption{MNIST-0.5, RL-attack}
    \end{subfigure}
    \caption{The performance of FL defense algorithms under different distribution conditions.}
    \label{fig:non_iid}
    
\end{figure*}

\subsection{Experimental Settings}

\subsubsection{Dataset}
We conduct experiments on four datasets: MNIST \cite{MNIST}, F-MNIST \cite{fashionmnist}, EMNIST \cite{emnist}, and Cifar10 \cite{cifar10}. Addressing the non-i.i.d. challenge in FL, we follow the approach from prior work \cite{LMP} by distributing training examples across all clients. Given an M-class dataset, clients are randomly divided into M groups. The probability \(q\) of assigning a training sample with label \(l\) to its respective group is set, with the probability of assigning it to other groups being \(\frac{1-q}{M-1}\). Training samples within the same client group adhere to the same distribution. When \(q=1/M\), the distribution of training samples across M groups is uniform, ensuring that all clients' datasets follow the same distribution. In cases where \(q>1/M\), the datasets among clients are not identically distributed. Using MNIST dataset, we set \(q=0.5\) to distribute training samples among clients unevenly, denoted as MNIST-0.5. MNIST-0.1 represents the scenario where MNIST is evenly distributed among clients (\(q=0.1\)).

\subsubsection{Metrics}
We assess FL defense methods by evaluating the global model's image classification accuracy after 500 epochs of FL training since these attacks aim to diminish testing accuracy. 
Higher accuracy of the global model under various attacks indicates stronger defense robustness.

\subsubsection{Baselines} 
To verify the effectiveness and stability of AdaAggRL, we mainly compare it with five other defense algorithms: Krum \cite{Krum}, coordinate-wise median (Median) \cite{Median}, norm clipping (Clipping) \cite{clipping}, an extension of the vanilla coordinate-wise median (C-Median) where a norm clipping step is applied \cite{RL-attack}, and FLtrust \cite{FLtrust}. Krum filters malicious updates at the client level, Clipping performs gradient clipping on parameter updates before aggregation, Median and C-Median select the median or clipped median of individual parameter values from all local model updates as global model parameters, and FLtrust requires the server to have access to an amount of root data. 
To further illustrate its performance, AdaAggRL is compared with Feddefender~\cite{Feddefender} and FedVal~\cite{fedval} on CIFAR-10 dataset.

We consider four poisoning attacks in FL: explicit boosting (EB) \cite{EB}, inner product manipulation (IPM) \cite{IPM}, local model poisoning attack (LMP) \cite{LMP}, and RL-based model attack (RL-attack) \cite{RL-attack}. IPM manipulates the attacker's gradients to ensure the inner product with true gradients becomes negative during aggregation. LMP generates malicious model updates by solving an optimization problem in each FL epoch. EB generates malicious updates through explicit enhancement, optimizing for a malicious objective designed to induce targeted misclassification. RL-attack adaptively generates attacks on the FL system using RL.

\subsubsection{Parameter Settings}
For MNIST, F-MNIST, and EMNIST, a Convolutional Neural Network (CNN) serves as the global model. In the case of Cifar10, the ResNet18 architecture \cite{resnet} is utilized as the global model.
In FL, there are 100 clients, denoted as $K=100$, with 20 malicious clients. 
For defense strategies based on RL, given the continuous action and state spaces, we select Twin Delayed DDPG (TD3) \cite{TD3} algorithm to train the defense policies in experiments\footnote{Source code: https://github.com/yjEugenia/AdaAggRL.}. Details on parameter determination are provided in the Appendix.

\subsection{Defense Performances}
\begin{table}[h!]
\begin{center}
\begin{footnotesize}
\begin{subtable}{0.98\columnwidth}
        \begin{tabular}{lcccc}
        \toprule
            & EB & IPM & LMP & RL-attack \\
        \midrule
        C-Median& 0.9598& 0.9537& 0.9329& 0.5550 \\
        Clipping        & 0.9151& 0.9654& 0.1944& 0.5750 \\
        FLtrust         & 0.9231& 0.9591& 0.9618& 0.7300 \\
        Krum            & 0.9325& 0.7897& 0.9458& 0.6875 \\
        Median          & 0.0981& 0.9549& 0.1135& 0.2750 \\
        AdaAggRL        & \textbf{0.9659}& \textbf{0.9658}& \textbf{0.9636}& \textbf{0.9655} \\
        \bottomrule
        \end{tabular}    
    \subcaption{CNN global model, MNIST-0.1}
    \label{tab3:subtab1}
\end{subtable}

\begin{subtable}{0.98\columnwidth}
\begin{tabular}{lcccc}
\toprule
    & EB & IPM & LMP & RL-attack \\
\midrule
C-Median& 0.9466& 0.9479& 0.9198& 0.4250\\
Clipping        & 0.7867& 0.9576& 0.0986& 0.4900\\
FLtrust         & 0.9488& 0.9414& 0.9412& 0.4375\\
Krum            & 0.9274& 0.7608& 0.9259& 0.1563\\
Median          & 0.0974& 0.9448& 0.1032& 0.1563\\
AdaAggRL        & \textbf{0.9608}& \textbf{0.9617}& \textbf{0.9604}& \textbf{0.9559}\\
\bottomrule
\end{tabular}
\subcaption{CNN global model, MNIST-0.5}
\label{tab3:subtab2}
\end{subtable}

\begin{subtable}{0.98\columnwidth}
\begin{tabular}{lcccc}
\toprule
    & EB & IPM & LMP & RL-attack \\
\midrule
C-Median& 0.7935& 0.8123& 0.7722& 0.6300 \\
Clipping        & 0.7249& 0.8261& 0.6015& 0.4600 \\
FLtrust         & 0.8154& 0.8344& 0.8386& 0.6150 \\
Krum            & 0.8082& 0.6610& 0.7942& 0.5625 \\
Median          & 0.1002& 0.8171& 0.1001& 0.0938 \\
AdaAggRL        & \textbf{0.8411}& \textbf{0.8398}& \textbf{0.8400}& \textbf{0.8337}  \\
\bottomrule
\end{tabular}
\subcaption{CNN global model, F-MNIST}
\label{tab3:subtab3}
\end{subtable}

\begin{subtable}{0.98\columnwidth}
\begin{tabular}{lcccc}
\toprule
    & EB & IPM & LMP & RL-attack \\
\midrule
C-Median& 0.8780& 0.8724& 0.8289& 0.1857 \\
Clipping        & 0.6694& \textbf{0.8834}& 0.0008& 0.1700 \\
FLtrust         & 0.8618& 0.8741& 0.8684& 0.2400 \\
Krum            & 0.8309& 0.5418& 0.8135& 0.0312 \\
Median          & 0.0388& 0.8716& 0.4331& 0.1850 \\
AdaAggRL        & \textbf{0.8816}& 0.8805& \textbf{0.8776}& \textbf{0.8786}  \\
\bottomrule
\end{tabular}
\subcaption{CNN global model, EMNIST}
\label{tab3:subtab4}
\end{subtable}

\begin{subtable}{0.98\columnwidth}
\begin{tabular}{lcccc}
\toprule
    & EB & IPM & LMP & RL-attack \\
\midrule
C-Median& 0.7091& 0.7364& 0.6048& 0.1150\\
Clipping        & 0.1002& 0.6589& 0.1388& 0.0850\\
FLtrust         & 0.5786& 0.6709& 0.7312& 0.6450\\
Krum            & 0.7238& 0.7028& 0.7283& 0.0900\\
Median          & 0.4484& 0.7309& 0.1018& 0.0900\\
AdaAggRL        & \textbf{0.7452}& \textbf{0.7497}& \textbf{0.7583}& \textbf{0.7531}\\
\bottomrule
\end{tabular}
\subcaption{ResNet18 global model, Cifar10}
\label{tab3:subtab5}
\end{subtable}
\caption{The testing accuracy of different FL aggregation methods under various attacks}
\label{result-table}
\end{footnotesize}
\end{center}
\end{table}
Table \ref{result-table} reports the testing accuracy of various FL aggregation methods across four datasets, indicating AdaAggRL's robustness. Across different datasets and models, AdaAggRL demonstrates consistent defensive efficacy against all four attack scenarios. 
Notably, AdaAggRL maintains stable effectiveness against RL-attack, where other defense methods face significant challenges and experience a noticeable degradation in defensive performance.
Figure \ref{fig:cifar10} illustrates the testing accuracy of the global model on Cifar10 dataset under four attacks, with different aggregation rules. AdaAggRL consistently outperforms other methods, achieving superior accuracy and convergence speed, particularly against RL-attack. Unlike FLtrust, it requires no root dataset and demonstrates more stable performance after 200 epochs, surpassing Krum and C-Median in all scenarios. 
AdaAggRL's performance on other datasets is shown in Appendix. 

LMP and EB attempt to optimize the objective function to make the poisoned gradients statistically inconspicuous, while RL-attack mimics the behavior of normal clients. As a result, defense methods like Median, C-Median, and Clipping, which rely solely on mean or median information through gradient clipping or selection, are prone to misjudge poisoned gradient updates. These methods may end up clipping correct updates while preserving erroneous ones.

AdaAggRL's performance is evaluated against Feddefender and FedVal on the CIFAR-10 dataset using ResNet18 over 100 FL epochs, shown in Table~\ref{table:newbaselines}. AdaAggRL is still significantly better than the latest baselines.
\begin{table}[tbp]
\begin{center}
\begin{footnotesize}
\begin{tabular}{lcccc}
\toprule
    & IPM & LMP & EB & RL-attack \\
\midrule
Feddefender            & 0.4711& 0.4196& 0.1008& 0.4150 \\
FedVal          & 0.6052& 0.0980& 0.4338& 0.1392 \\
AdaAggRL        & \textbf{0.6984}& \textbf{0.7063}& \textbf{0.7143}& \textbf{0.7106}  \\
\bottomrule
\end{tabular}
\end{footnotesize}
\end{center}
\caption{The accuracy of aggregation methods under attacks}
\label{table:newbaselines}
\end{table}
\subsection{Ablation Studies}

\subsubsection{Impact of Current-history Similarity}
To illustrate the impact of the similarity metrics $S_{cl}$ on the stability of the FL process, Figure \ref{Scl} depicts the defense performance of AdaAggRL compared to the case where $S_{cl}$ is not considered under LMP attack. We observe that considering the variations in $S_{cl}$ indeed enhances the convergence speed and reduces the oscillation amplitude of testing accuracy.
\subsubsection{Impact of Current (history)-global Similarity}
Figure \ref{Scg} illustrates the defense performance of AdaAggRL compared to the case where $S_{cg}$ and $S_{lg}$ are not considered under EB attack. We observe that the malicious client data distribution obtained through gradient reversal may stably deviate from the normal distribution, leading to an inflated $S_{cl}$. If $S_{cg}$ and $S_{lg}$ are not considered, the defense effectiveness degrades.
\subsection{Analysis}
\subsubsection{Impact of the Number of Attackers}
Figures \ref{attacker_lmp} and \ref{attacker_eb} show testing accuracy under LMP and EB attacks as malicious client proportion increases from 0\% to 95\%. Both AdaAggRL and FLtrust can tolerate up to 90\% of malicious clients. AdaAggRL shows a slight accuracy decline as malicious client proportions increase, while the remaining FL aggregation algorithms can only tolerate malicious clients below 30\% under LMP attack. This highlights AdaAggRL's stability against high percentages of malicious clients.

\subsubsection{Impact of non-i.i.d. Degree}
Table \ref{tab3:subtab2} reports testing accuracy of FL defense algorithms on MNIST-0.5 under various attacks, while Figure \ref{fig:non_iid} compares their performance under different distribution conditions. With non-i.i.d. data (q=0.5), baselines show reduced accuracy under LMP and increased oscillations under EB, and non-i.i.d. conditions significantly impact their performance against RL-attack.  AdaAggRL demonstrates faster convergence, particularly against IPM, LMP, and EB, with minimal performance decline across attacks, demonstrating robustness to non-i.i.d. data. Under significant non-i.i.d. impact, RL can adaptively lower the weights of global model-related similarity scores. More results for q$>$0.5 can be found in the Appendix.

\section{Conclusion}
In this paper, we propose AdaAggRL, an RL-based Adaptive Aggregation method, to counter sophisticated poisoning attacks. Specifically, we first utilize distribution learning to simulate clients' data distributions. Then, we use MMD to calculate the pairwise similarity of the current local model data distribution, its historical data distribution, and the global model data distribution. Finally, we use policy learning to adaptively determine the aggregation weights based on the above similarities and the reconstruction similarity.
Experiments on four real-world datasets demonstrate that AdaAggRL significantly outperforms the state-of-the-art defense model for sophisticated attacks. 
Future work can involve investigating novel methods to attack the adaptive defense approach. One potential scheme to construct malicious update parameters is to solve an optimization problem under the condition of controlling the stability of simulated data distribution changes.

\section{Acknowledgments}
This work was funded by the National Natural Science Foundation of China (NSFC) under Grants No. 62406013, the  Beijing Advanced Innovation Center Funds for Future Blockchain and Privacy Computing(GJJ-23-006) and the Fundamental Research Funds for the Central Universities. 

\bibliography{aaai25}
\newpage
\appendix
\section{Method pseudocode} \label{appendix:code}
\begin{algorithm}[htbp]
   \caption{AdaAggRL}
   \label{alg:AdaAggRL}
\begin{algorithmic}
   \STATE {\bfseries Input:} $K$ clients; learning rate $\alpha$; number of global iterations $R_g$; number of local iterations $R_l$
   \STATE {\bfseries Output:} Global model $\theta$
   \STATE $\theta^0 \gets$random initialization.
   \FOR{$t=0, 1,\dots, R_g$}
   \STATE // Transition of environmental state
   \STATE The server randomly samples clients subset $C^t$ and sends $\theta^t$ to them.
   \FOR{$k \in C^t$}
   \STATE $\theta^{t+1}_k \gets  
   \begin{cases} 
      \theta^t-\alpha\nabla F_k
      (\theta)& \text{Benign clients} \\
      * & \text{Malicious clients} 
    \end{cases}$
    \STATE send $\theta^{t+1}_k$ to the server
   \ENDFOR
   \STATE $\mathbf{s}^t \gets EnvironmentalCues(\theta^t,\theta^{t+1}_{C^t})$
   \STATE // Obtain actions
   \STATE $\mathbf{A}^t=(\mathbf{a}^t,b^t)\gets Actions(\mathbf{s}^t)$
   \STATE $\tilde{\mathbf{w}}\gets g(\mathbf{s}^t\cdot\mathbf{a}^t)$
   \STATE // Calculate the reward
   \STATE $\delta\gets \max(\tilde{\mathbf{w}})\cdot b^t$
   \STATE $\mathbf{w} = f_{\delta}(\tilde{\mathbf{w}})$
   \STATE $\theta^{t+1}\gets \Sigma_{k\in C^t}w_k/\lambda^{h^{t+1}_k}\cdot\theta^{t+1}_k$
   \COMMENT {$\lambda^{h^{t+1}_k}$ indicates the penalty for malicious behavior.}
   \STATE $reward\gets f(\theta^t) - f(\theta^{t+1})$
   \ENDFOR\\
   \STATE {\bfseries return} $\theta = \theta^{t+1}$
\end{algorithmic}
\end{algorithm}

\begin{algorithm}[!h]
   \caption{EnvironmentalCues}
   \label{alg:environment}
\begin{algorithmic}
   \STATE {\bfseries Input:} Globel model parameters $\theta^t$; The set of client model parameters participating in the training $\theta_{C^t}$
   \STATE {\bfseries Output:} Environment state $\mathbf{s}^t$
   \FOR{$k \in C^t$}
   \STATE $\bar{g}^t_k \gets (\theta^{t+1}_k-\theta^{t})/\alpha$
   \STATE $D_{rec,k},S^t_{k,R}\gets \mathrm{IG}(\bar{g}^t_k)$
   \STATE {$V^{history}_k\gets V^{current}_k$}
   \STATE {$V^{current}_k\gets \mathrm{CNN}(D_{rec,k})$}
   \ENDFOR
   \STATE $V_g \gets \mathrm{average}(\{V^{current}_k\})$
   \FOR{$k \in C^t$}
   \STATE $S_{k,cl}\gets 2\cdot\cos(\tanh(\mathrm{MMD}(V^{current}_k,V^{history}_k)/2))-1$
   \STATE $S_{k,cg}\gets 2\cdot\cos(\tanh(\mathrm{MMD}(V^{current}_k,V_g)/2))-1$
   \STATE $S_{k,lg}\gets 2\cdot\cos(\tanh(\mathrm{MMD}(V^{history}_k,V_g)/2))-1$
   \STATE $s^t_k\gets (S_{k,R}, S_{k,cl}, S_{k,cg}, S_{k,lg})$
   \ENDFOR
   \STATE $\mathbf{s}^t \gets (s^t_{k_1}, s^t_{k_2}, ..., s^t_{k_{|C^t|}})^T$\\
   \STATE {\bfseries return} $\mathbf{s}^t$
\end{algorithmic}
\end{algorithm}
Algorithm \ref{alg:AdaAggRL} presents the comprehensive AdaAggRL approach developed in our study. AdaAggRL is reinforcement learning-based, where the Federated Learning (FL) system undergoes three processes in each epoch: environmental state transition, acquiring actions based on the policy model, and computing rewards for the current step. The calculation of the environmental state, based on the model parameters uploaded by clients, is accomplished through the $\mathrm{EnvironmentalCues}$ method outlined in Algorithm \ref{alg:environment}. The $\mathrm{Actions}$ function utilizes reinforcement learning to obtain the policy based on the current environmental state.

In the $\mathrm{EnvironmentalCues}$ method of Algorithm \ref{alg:environment}, $\mathrm{IG}$ denotes the application of the gradient reversal technique for reshaping the dataset distribution and computing the reconstruction similarity. $\mathrm{CNN}$ represents a convolutional neural network employed for extracting image features, while $\mathrm{MMD}$ indicates the utilization of Maximum Mean Discrepancy (MMD) to quantify the distribution disparity between features.

\section{Details of Experimental Settings}
The experiments are conducted on one Nvidia A100 40G GPU. We implement all models by PyTorch \cite{PyTorch}. In FL, there are 100 clients, denoted as $K=100$, with 20 malicious clients. It trains for 500 epochs, wherein each epoch, the server randomly selects 10\% of the clients to participate. Each client conducts one round of training on its local dataset with a learning rate of 0.05.
For defense strategies based on RL, given the continuous action and state spaces, we select Twin Delayed DDPG (TD3) \cite{TD3} algorithm to train the defense policies in experiments. 
We configure parameters such that the policy model uses 'MlpPolicy', the learning rate is set to \(1 \times 10^{-5}\) for the Adam optimizer, the batch size is 64 for each gradient update, and the discount factor \(\gamma\) is set to 0.99. We then present the results obtained using TD3. We maintain fixed initial models and random seeds for data sampling to ensure a fair comparison.

In training setup, the RL attack model first learns AD hoc attack strategies against other baseline defense methods, and then fixes the parameters. In response, we developed a temporary defense strategy to defend against this learned attack strategy. On this basis, the RL attack model refines the attack strategy against RL defense according to our AD hoc defense strategy. We then implement defensive measures based on this improved attack strategy and evaluate the effectiveness of our defense mechanisms. Despite the attacker's awareness of the server's utilization of RL-based defense, our algorithm remains capable of detecting fluctuations in its simulated data distribution, thus identifying malicious behavior and diminishing the attacker's involvement in FL training, ultimately leading to their withdrawal from the training process.

In each round of FL training, we reconstruct 16 images from a single model update of a sampled client to simulate data distribution. The optimization process for gradient inversion is carried out for 30 iterations. However, it's important to note that the reconstructed images are solely used to capture the distribution characteristics and are not intended for precise image reconstruction. The limited number of iterations and reconstructed images do not suffice for accurately restoring image information. Consequently, the reconstructed images do not contain discernible image details. This approach is implemented to fulfill the privacy protection requirements of FL while minimizing computational time as much as possible. The experimental results demonstrate that, under the LMP attack, when reconstructing 16 images from the MNIST dataset, the final accuracy achieves after 10 optimization iterations of gradient inversion is 0.9554, after 20 iterations is 0.9592, after 30 iterations is 0.9636, and after 50 iterations is 0.9664. When reconstructing 32 images, the final accuracy after 30 optimization iterations is 0.9628. AdaAggRL defense performance is weakly correlated with max\_iters and num\_images, not leading to unbounded computational overhead. Increasing max\_iters to 50 only slightly improves accuracy over max\_iters=10 (by about 1.15\%), and num\_images=16 achieves comparable accuracy to num\_images=32.

To strike a balance between the final performance and computational resources, we opt to train using gradient inversion with 30 iterations for reconstructing 16 images. The method for gradient inversion optimization is Adam, the learning rate is 0.05, and the reconstructed images are initialized to all-zero tensors. And the fixed parameter $\beta$ in the optimization target is set to 1e-4.
\section{Supplementary Experimental Results} \label{appendix:exp}
\begin{figure}[htbp]
    \centering
    \begin{subfigure}{0.43\linewidth}
        \includegraphics[width=1\linewidth]{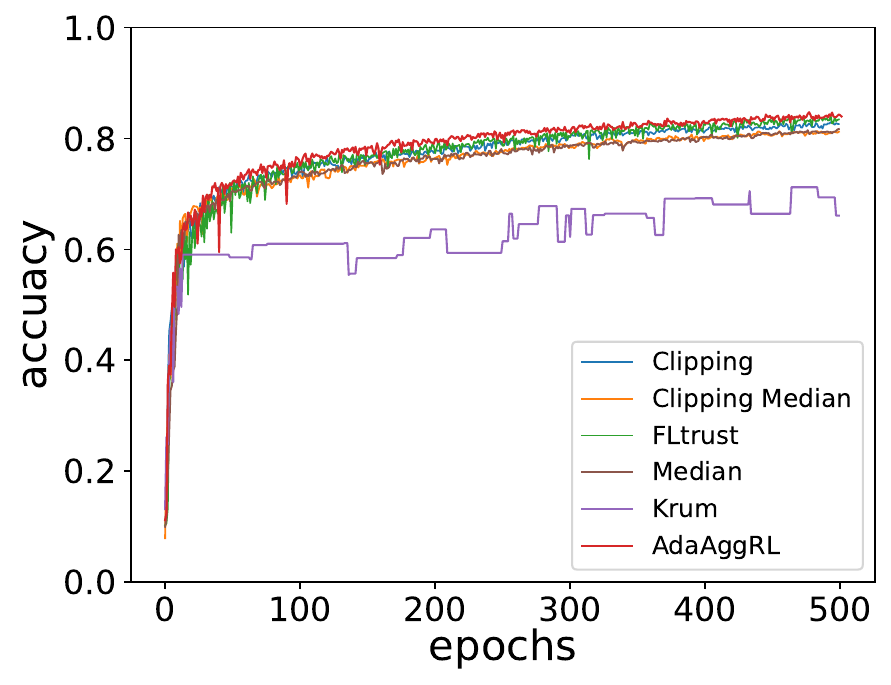}
        \subcaption{IPM}
    \end{subfigure}
    \begin{subfigure}{0.43\linewidth}
        \includegraphics[width=1\linewidth]{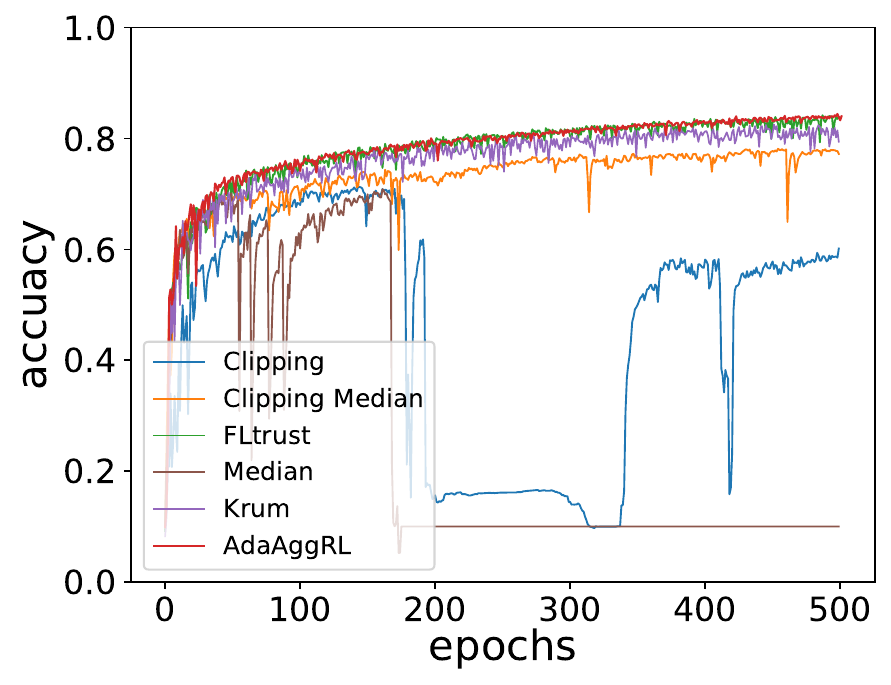}
        \subcaption{LMP}
    \end{subfigure}
    \begin{subfigure}{0.43\linewidth}
        \includegraphics[width=1\linewidth]{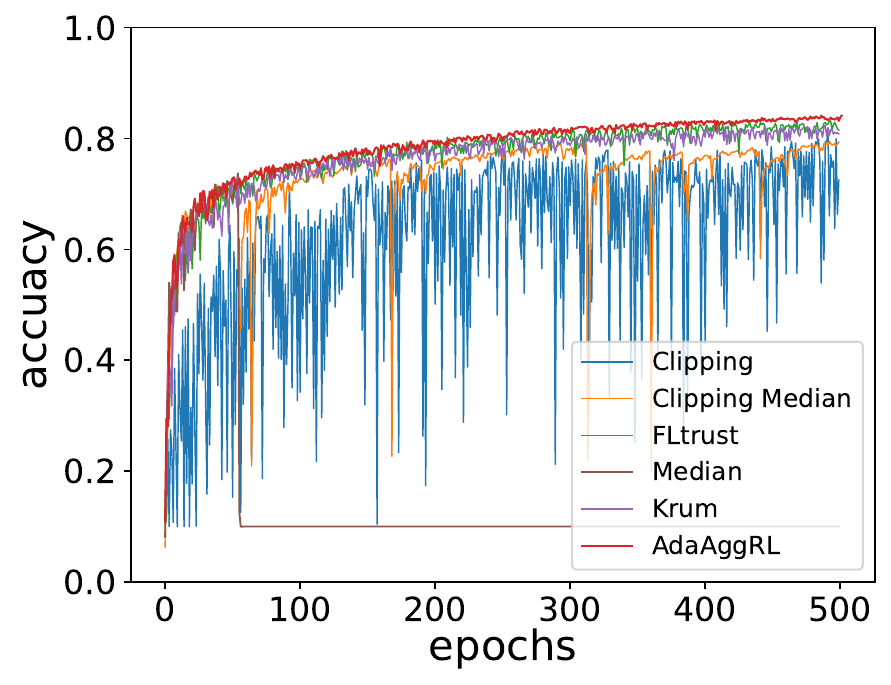}
        \subcaption{EB}
    \end{subfigure}
    \begin{subfigure}{0.43\linewidth}
        \includegraphics[width=1\linewidth]{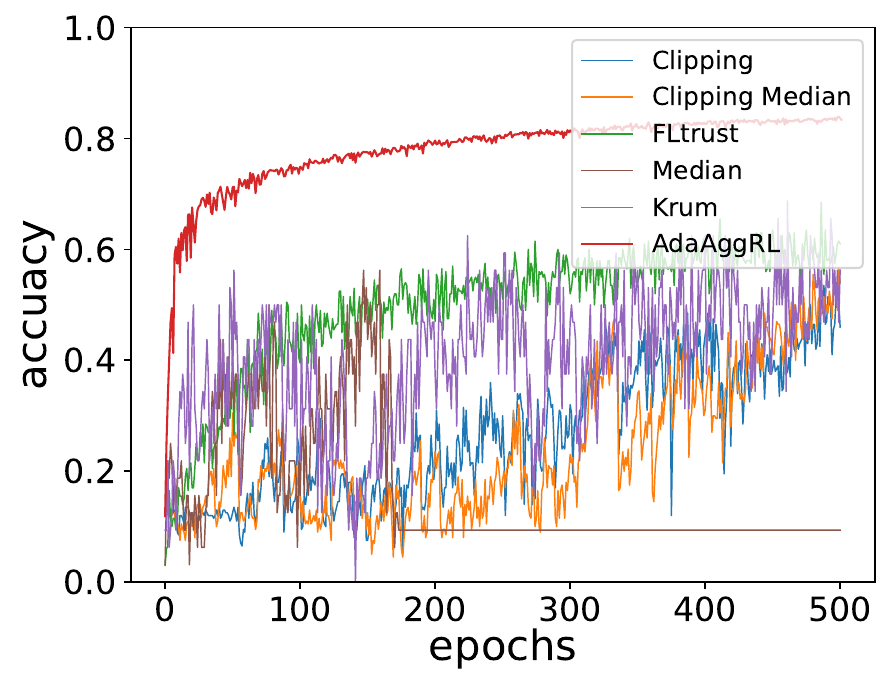}
        \subcaption{RL-attack}
    \end{subfigure}
    \caption{The variation in the testing accuracy of the global model over FL epochs on F-MNIST dataset considering different attacks.}
    \label{fig:fashion}
    
\end{figure}

Figure \ref{fig:fashion} illustrates the test accuracy variations over FL epochs for different FL aggregation methods under four attacks (IPM, LMP, EB, and RL-attack) on F-MNIST dataset. We observe that AdaAggRL achieves higher accuracy and more stable convergence, especially when facing the more sophisticated RL-attack.

\begin{figure}[tbp]
    \centering
    \begin{subfigure}{0.43\linewidth}
        \includegraphics[width=1\linewidth]{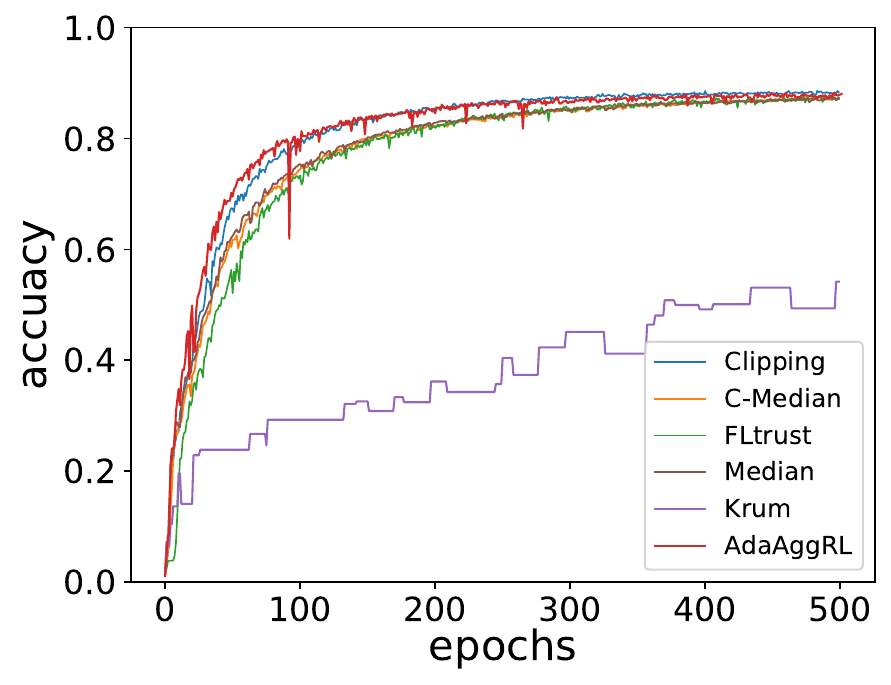}
        \subcaption{IPM}
    \end{subfigure}
    \begin{subfigure}{0.43\linewidth}
        \includegraphics[width=1\linewidth]{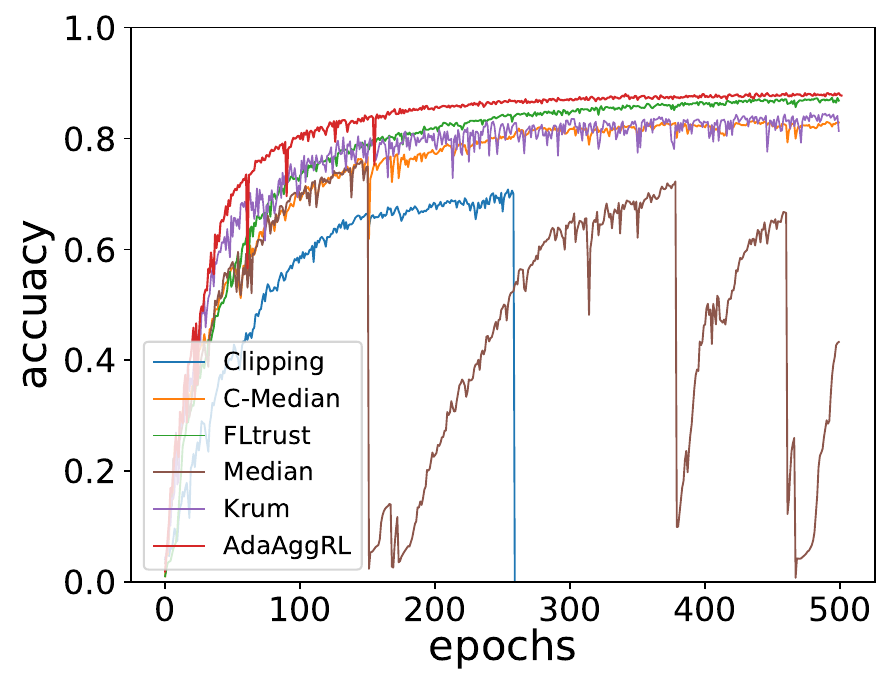}
        \subcaption{LMP}
    \end{subfigure}
    \begin{subfigure}{0.43\linewidth}
        \includegraphics[width=1\linewidth]{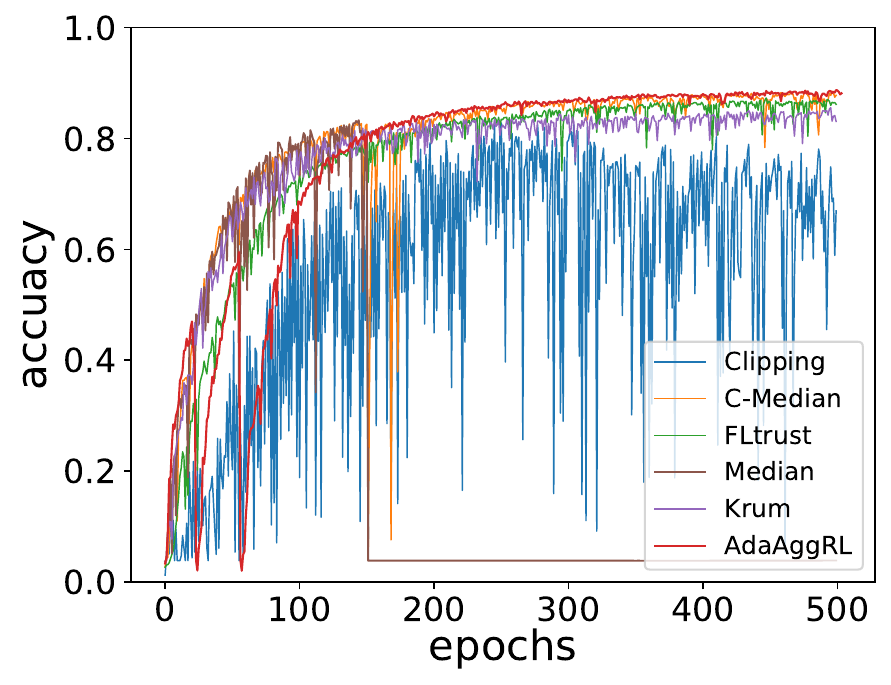}
        \subcaption{EB}
    \end{subfigure}
    \begin{subfigure}{0.43\linewidth}
        \includegraphics[width=1\linewidth]{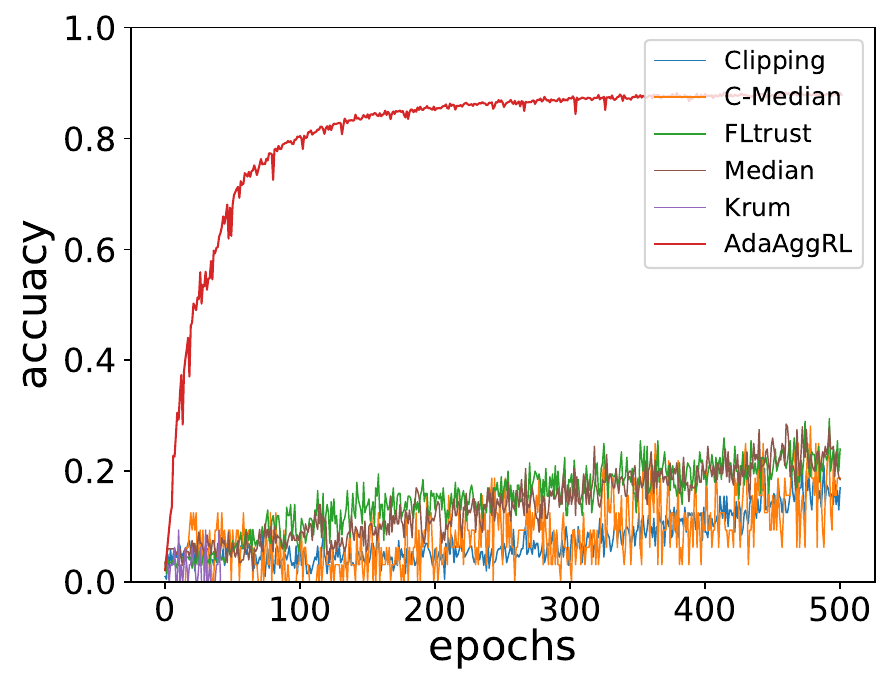}
        \subcaption{RL-attack}
    \end{subfigure}
    \caption{The variation in the testing accuracy of the global model over FL epochs on EMNIST dataset considering different attacks.}
    \label{fig:emnist}    
\end{figure}

Figure \ref{fig:emnist} illustrates the test accuracy variations over FL epochs for different FL aggregation methods under four attacks (IPM, LMP, EB, and RL-attack) on EMNIST dataset. We observe that AdaAggRL exhibits faster convergence under IPM, achieves higher accuracy under LMP and EB, and demonstrates particularly stable training outcomes and increased accuracy when faced with RL-attack.

Figure \ref{fig:emnist} illustrates the test accuracy variations over FL epochs for different FL aggregation methods under four attacks (IPM, LMP, EB, and RL-attack) on MNIST-0.9 (q=0.9) dataset. We observe that in the case of more extreme heterogeneity of data distribution, AdaAggRL has higher convergence speed and accuracy, and has a stronger ability to re-resume training.
\begin{figure}[tbp]
    \centering
    \begin{subfigure}{0.43\linewidth}
        \includegraphics[width=1\linewidth]{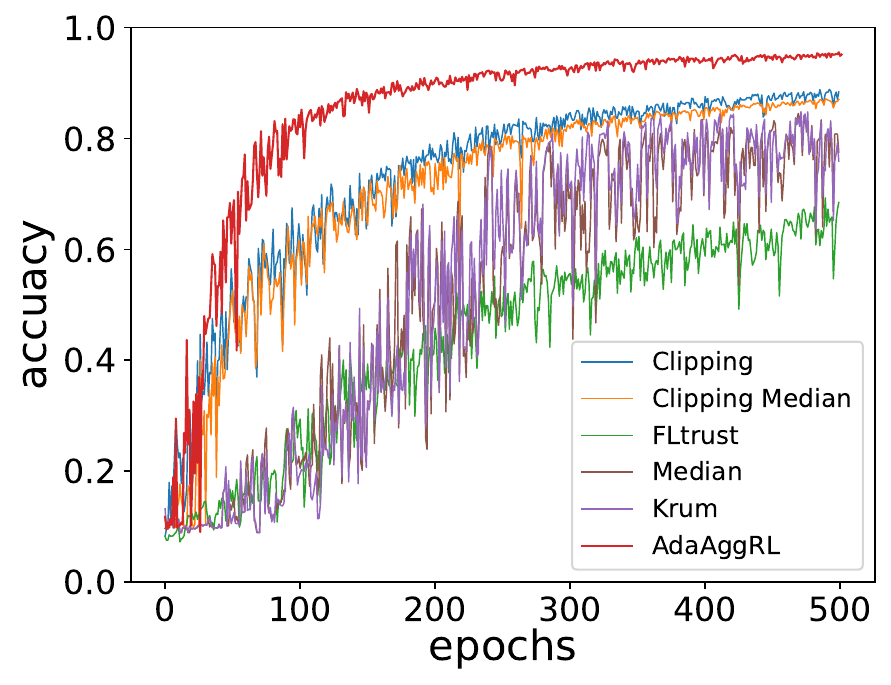}
        \subcaption{IPM}
    \end{subfigure}
    \begin{subfigure}{0.43\linewidth}
        \includegraphics[width=1\linewidth]{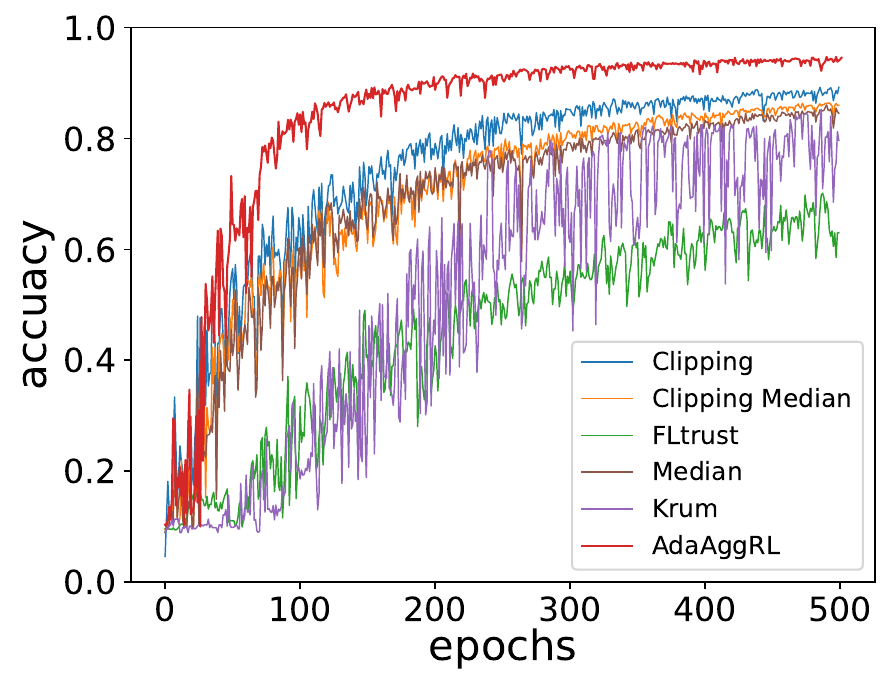}
        \subcaption{LMP}
    \end{subfigure}
    \begin{subfigure}{0.43\linewidth}
        \includegraphics[width=1\linewidth]{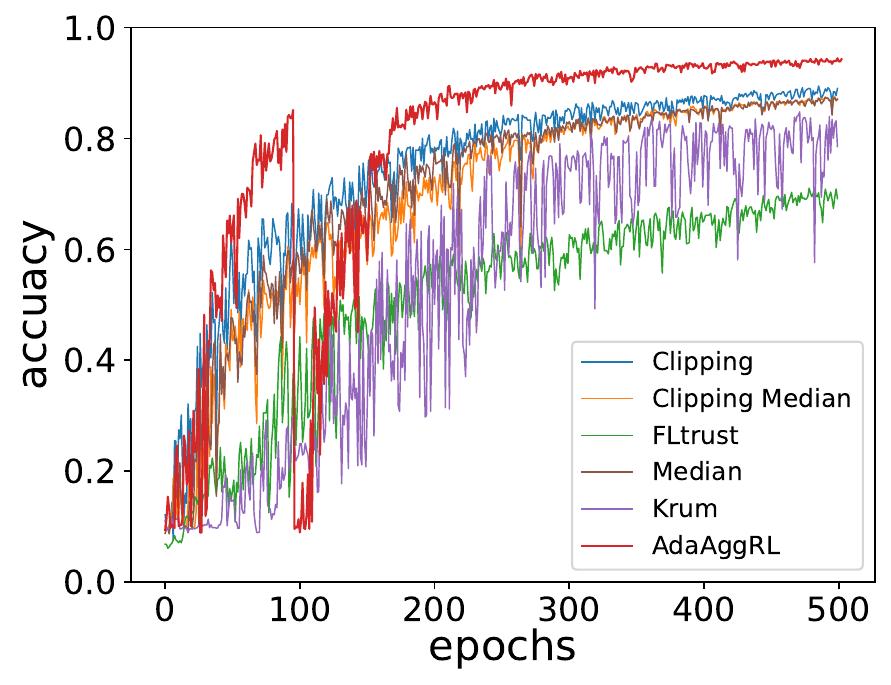}
        \subcaption{EB}
    \end{subfigure}
    \begin{subfigure}{0.43\linewidth}
        \includegraphics[width=1\linewidth]{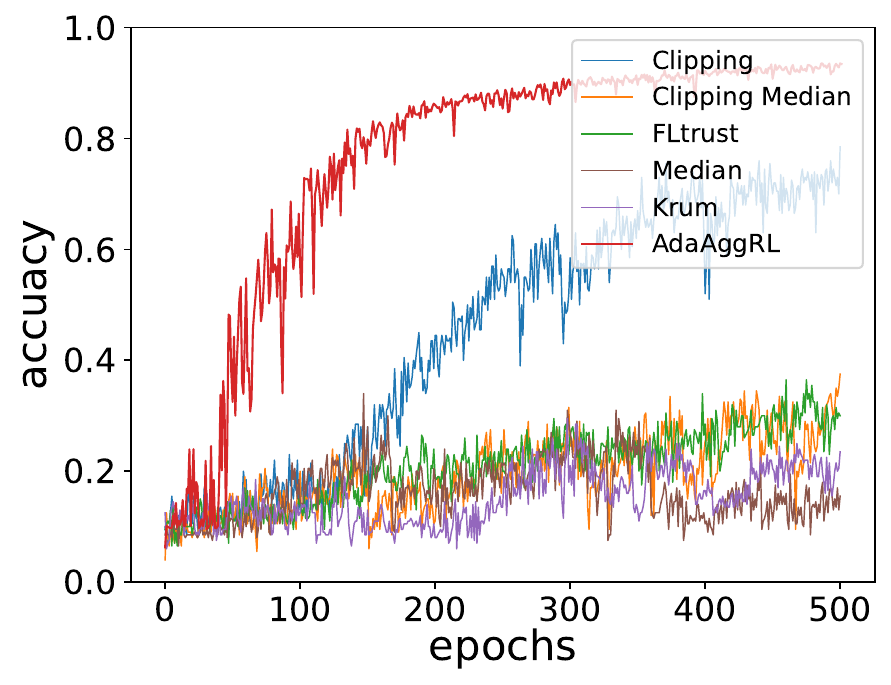}
        \subcaption{RL-attack}
    \end{subfigure}
    \caption{The variation in the testing accuracy of the global model over FL epochs on MNIST-0.9 dataset considering different attacks.}
    \label{fig:mnist0.9}    
\end{figure}

\begin{figure}[htbp]
    \centering
    \begin{subfigure}{0.43\linewidth}
        \includegraphics[width=1\linewidth]{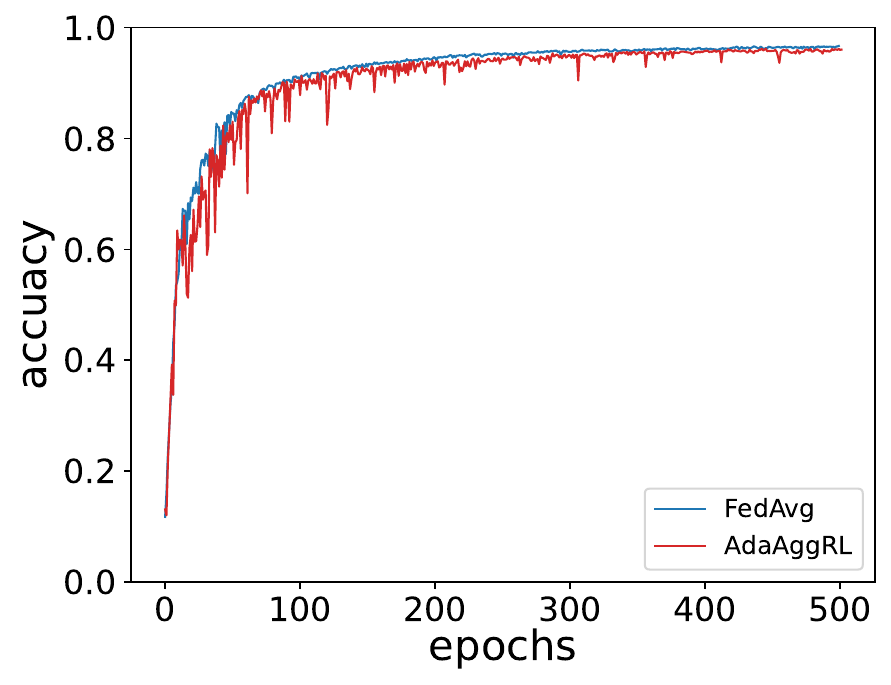}
        \subcaption{MNIST-0.1}
    \end{subfigure}
    \begin{subfigure}{0.43\linewidth}
        \includegraphics[width=1\linewidth]{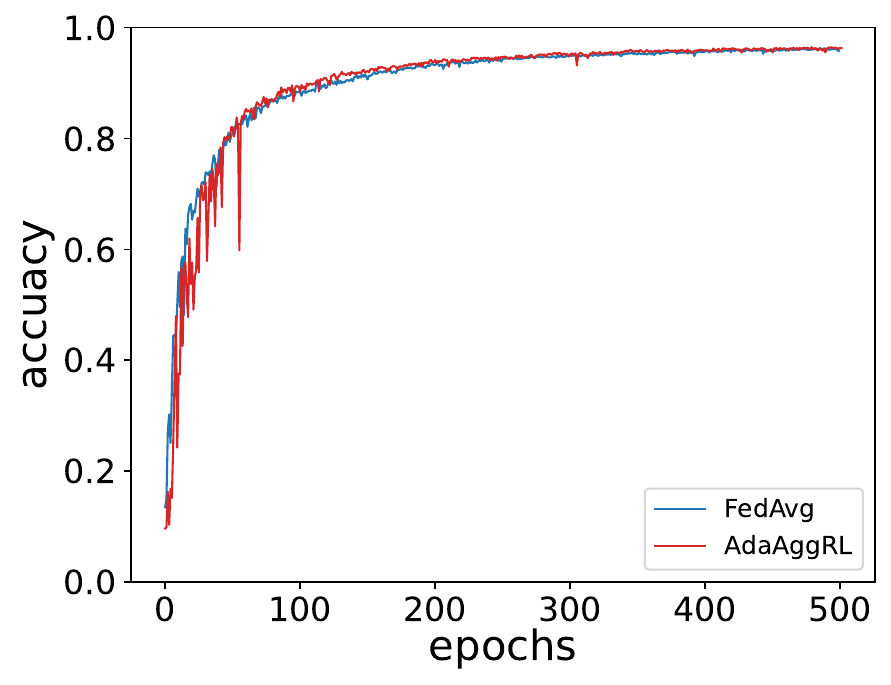}
        \subcaption{MNIST-0.5}
    \end{subfigure}
    \begin{subfigure}{0.43\linewidth}
        \includegraphics[width=1\linewidth]{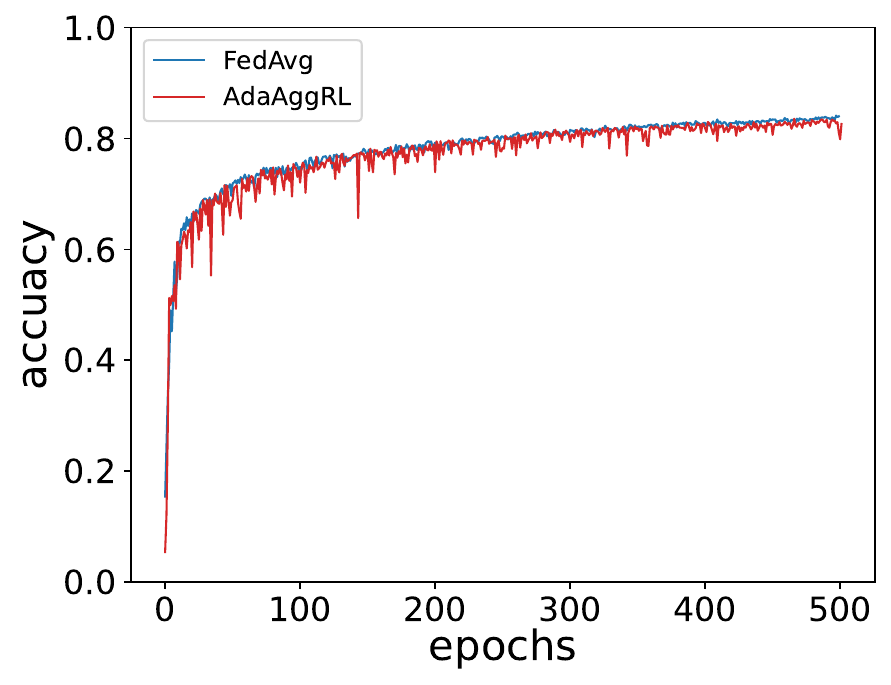}
        \subcaption{F-MNIST}
    \end{subfigure}
    \caption{The comparison of test accuracy between FedAvg and AdaAggRL under attack-free conditions on different datasets.}
    \label{fig:fedavg}
    
\end{figure}

Figure \ref{fig:fedavg} illustrates the test accuracy variations over FL epochs for AdaAggRL and FedAvg under no attacks. We observe that in the absence of attacks, AdaAggFL and FedAvg perform similarly on MNIST and F-MNIST datasets. However, in the presence of non-i.i.d. data (MNIST-0.5), AdaAggRL achieves a slightly higher test accuracy than FedAvg after 500 rounds of training.

\begin{figure}[htbp]
    \centering
    \begin{subfigure}{0.43\linewidth}
        \includegraphics[width=1\linewidth]{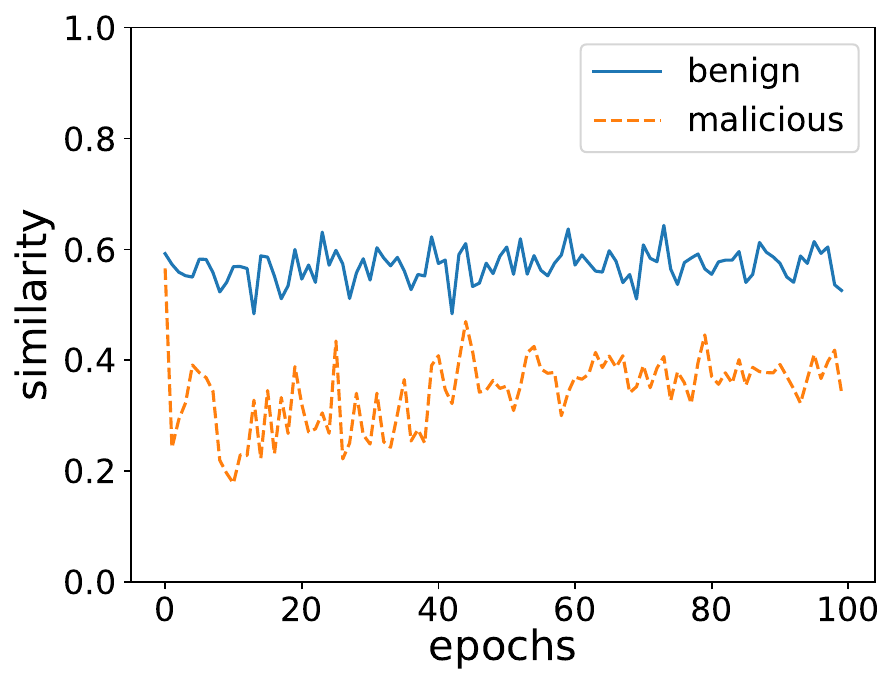}
        \subcaption{IPM}
    \end{subfigure}
    \begin{subfigure}{0.43\linewidth}
        \includegraphics[width=1\linewidth]{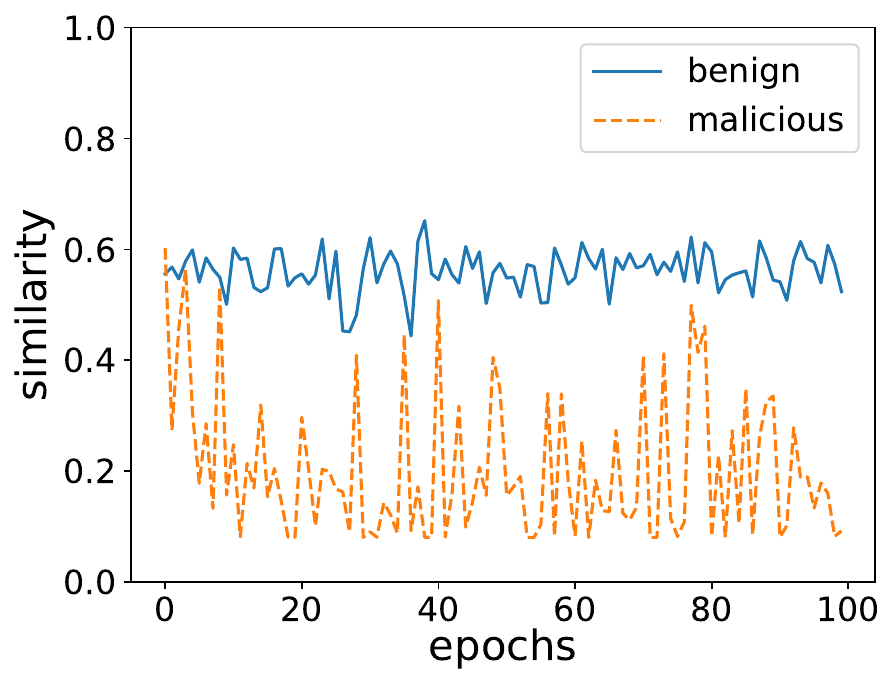}
        \subcaption{LMP}
    \end{subfigure}
    \begin{subfigure}{0.43\linewidth}
        \includegraphics[width=1\linewidth]{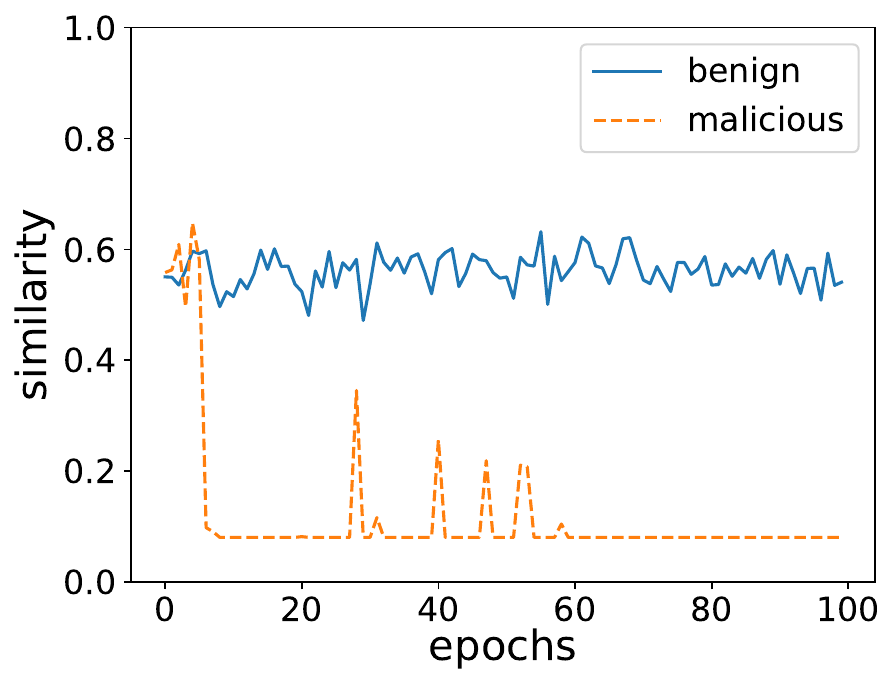}
        \subcaption{EB}
    \end{subfigure}
    \begin{subfigure}{0.43\linewidth}
        \includegraphics[width=1\linewidth]{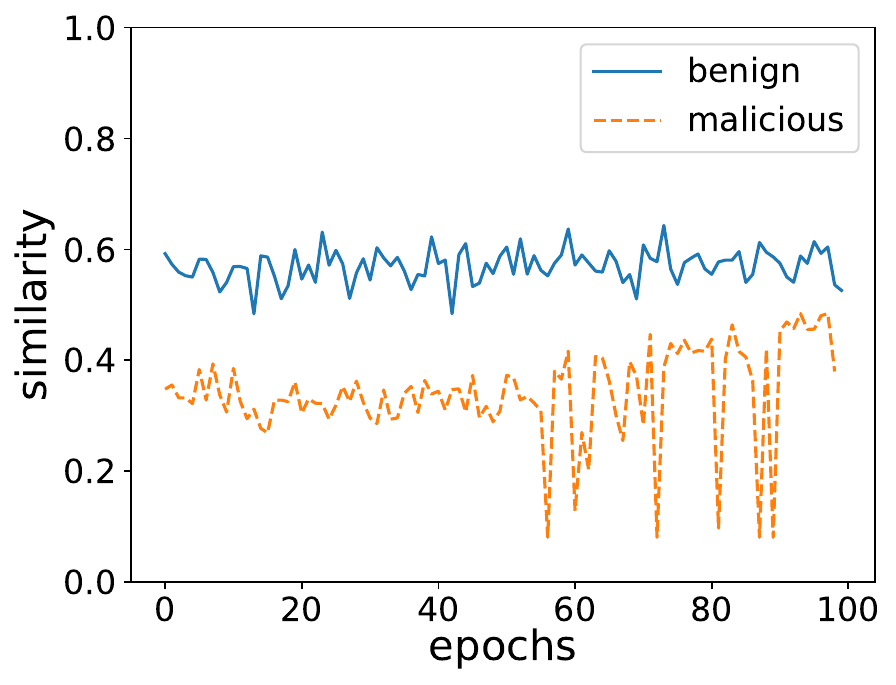}
        \subcaption{RL-attack}
    \end{subfigure}
    \caption{The statistical results of the similarity between the historical client data distribution and the global model data distribution under four types of attacks vary with the training epochs on MNIST dataset.}
   
    \label{fig:lg}
\end{figure}

We observe a certain degree of similarity between the historical client data distribution and the global model data distribution, as shown in Figure~\ref{fig:lg}. We find that the data distribution of benign clients for historical steps is similar to the global model and remains stable, but the similarity of malicious clients is lower. This is because the data distribution of the global model represents the average state of normal data, so the distribution of benign clients should always be consistent with the global model. Instead, malicious clients do not possess this property. Though the advanced RL-attack has a high similarity with the global model due to collecting the data distribution of the server, it is still not as stable as the benign client. While considering the similarity between the current client data distribution and the global model data distribution, we also take into account the similarity between historical client data distribution and global model data distribution. By incorporating the historical performance of clients, this approach aims to capture the subtle behaviors of lurking malicious clients engaging in attacks.

\begin{figure}[htbp]
    \centering
    \begin{subfigure}{0.43\linewidth}
    \includegraphics[width=1\linewidth]{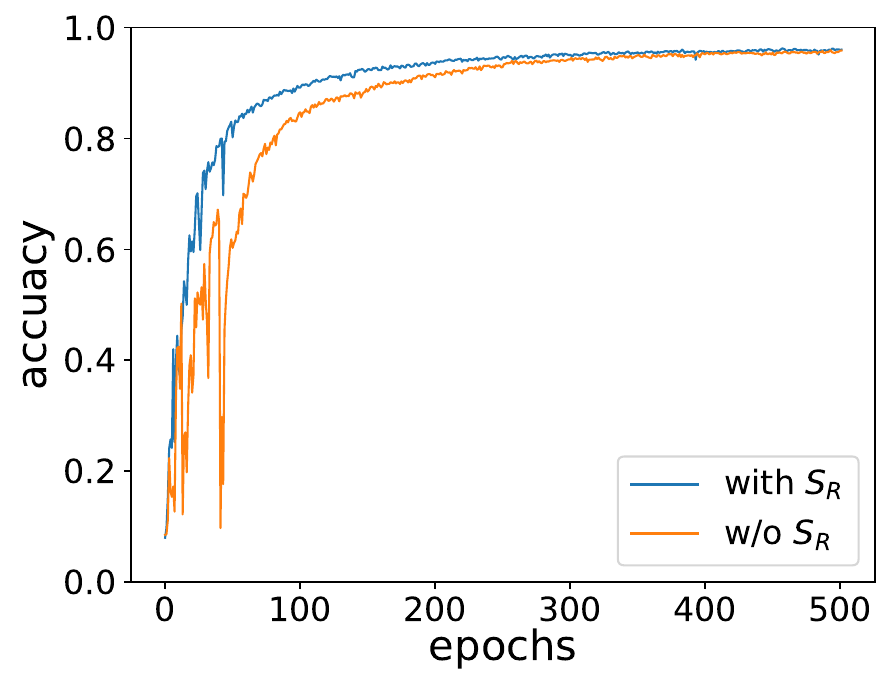}
    \subcaption{LMP}
    \end{subfigure}
    \caption{The defense performance of AdaAggRL compared to the case where $S_R$ is not considered under LMP attack }
    \label{fig:Sr}
\end{figure}

To illustrate the impact of the similarity metrics $S_{R}$ between the current client data distribution and historical data distribution on the stability of the federated learning process, Figure \ref{fig:Sr} depicts the defense performance of AdaAggRL compared to the case where $S_{R}$ is not considered under LMP attack. We observe that considering the variations in $S_{R}$ indeed enhances the convergence speed and reduces the oscillation amplitude of testing accuracy.
\section{Comparison with Advanced Baselines}
To demonstrate the effectiveness of AdaAggRL, we also compare it with more novel baselines, Feddefender~\cite{Feddefender} and FedVal~\cite{fedval}, on the CIFAR-10 dataset. Using ResNet18 as the base model and a learning rate of 0.01, we conducted FL for 100 epochs. We records the accuracy, as shown in the table~\ref{table:newbaselines}. We can see that our AdaAggRL method is still significantly better than the latest proposed  baselines.
\begin{table}[htbp]
\caption{The testing accuracy of different FL aggregation methods under various attacks}
\label{table:newbaselines}
\begin{center}
\begin{scriptsize}
\begin{tabular}{lcccc}
\toprule
    & IPM & LMP & EB & RL-attack \\
\midrule
Feddefender            & 0.4711& 0.4196& 0.1008& 0.4150 \\
FedVal          & 0.6052& 0.0980& 0.4338& 0.1392 \\
AdaAggRL        & \textbf{0.6984}& \textbf{0.7063}& \textbf{0.7143}& \textbf{0.7106}  \\
\bottomrule
\end{tabular}
\end{scriptsize}
\end{center}
\end{table}
\section{NLP Dataset Distribution Changes Caused by Poison Attacks} \label{appendix:nlp}
\begin{figure}[tbp]
    \centering
    \begin{subfigure}{0.43\linewidth}
        \includegraphics[width=1\linewidth]{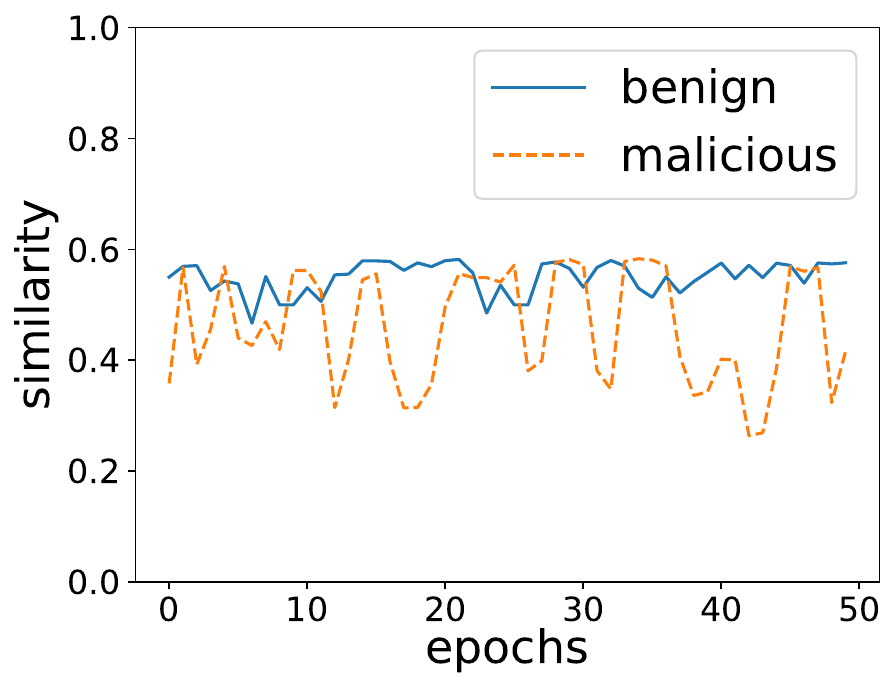}
        \subcaption{IPM}
    \end{subfigure}
    \begin{subfigure}{0.43\linewidth}
        \includegraphics[width=1\linewidth]{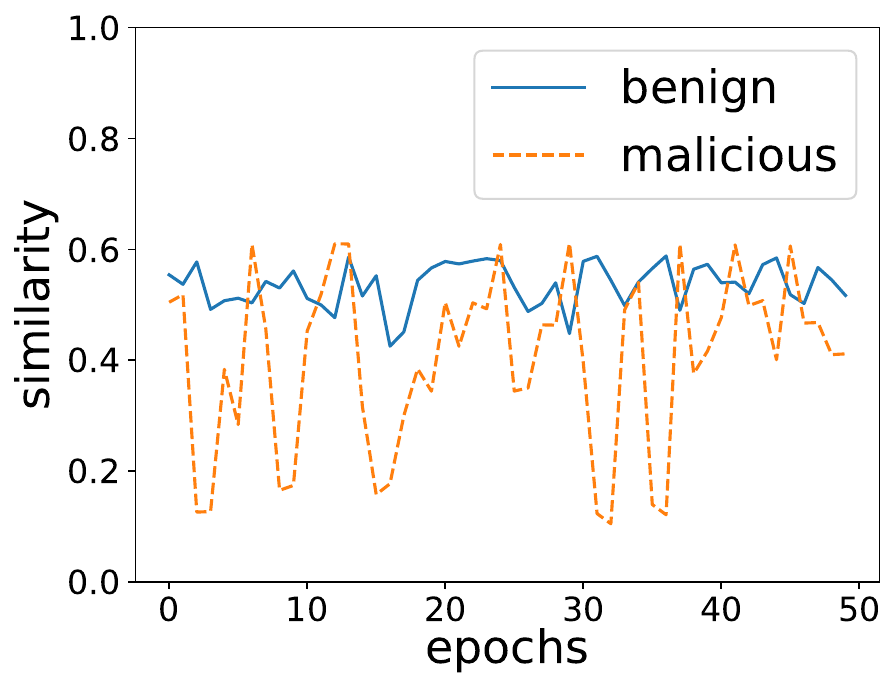}
        \subcaption{LMP}
    \end{subfigure}
    \begin{subfigure}{0.43\linewidth}
        \includegraphics[width=1\linewidth]{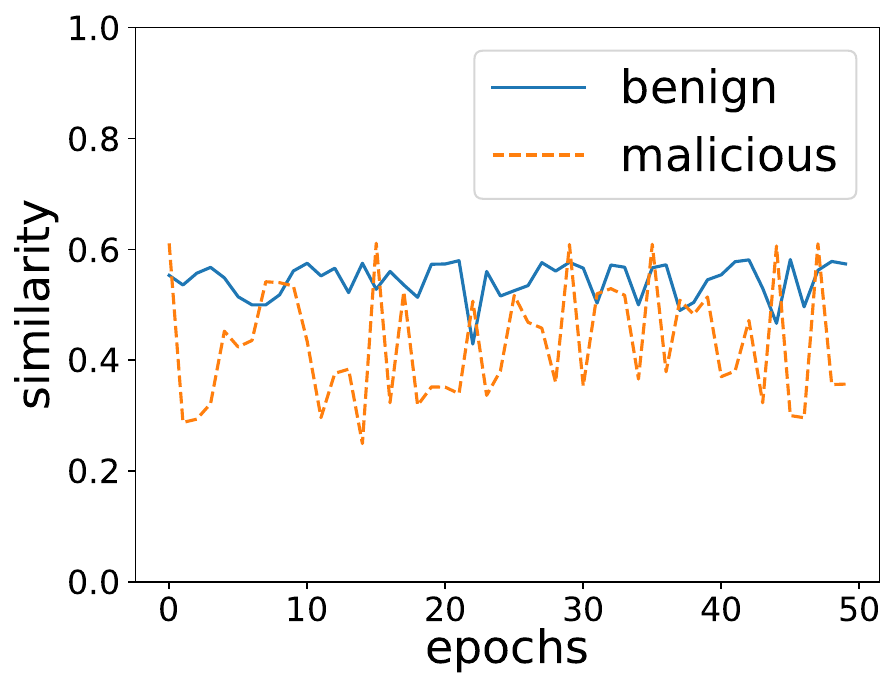}
        \subcaption{EB}
    \end{subfigure}
    \begin{subfigure}{0.43\linewidth}
        \includegraphics[width=1\linewidth]{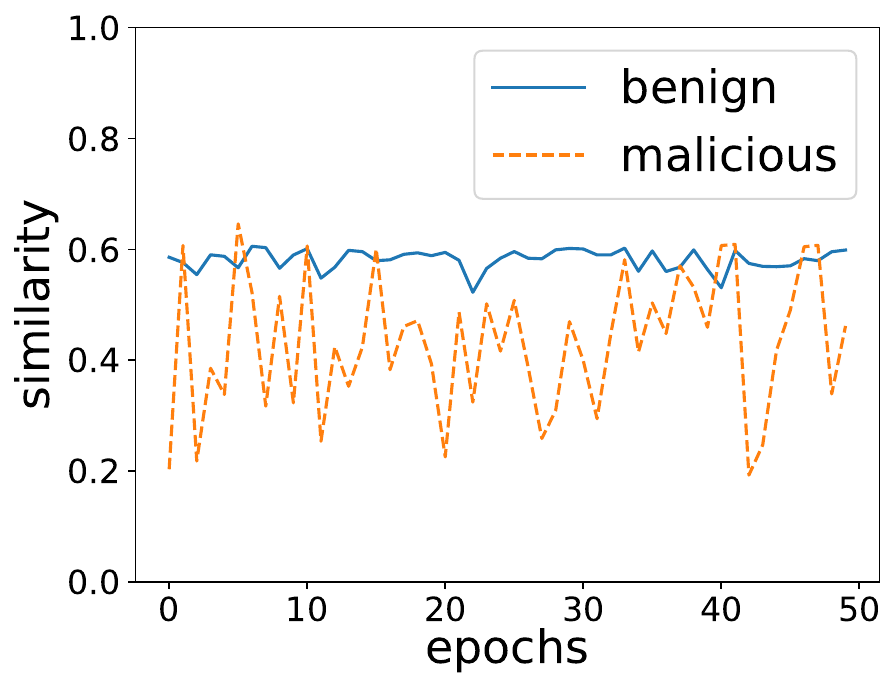}
        \subcaption{RL-attack}
    \end{subfigure}
    \caption{The statistical results of the similarity between the current client data distribution and its historical data distributions under four types of attacks that vary with the training epochs.}
    \label{fig:nlp}
\end{figure}
In order to verify whether the hypothesis that poisoning attacks can cause abnormal changes in the distribution of datasets reconstructed by gradient inversion is valid on NLP datasets, we select IMDB review dataset~\cite{IMDB} on TextCNN~\cite{textcnn} framework for FL semantic classification. IPM, LMP, EB and RL-attack poisoning attacks are also carried out during training. Unlike the image data set, we record and observe changes in the distribution of embedding. Figure~\ref{fig:nlp} shows the statistical results of the similarity between the current client data distribution and its historical data distributions under four types of attacks that vary with the training epochs. 
It can be seen that on IMDB dataset, under different poisoning attacks, the similarity between the historical embedding distribution and the current embedding distribution of malicious clients is generally lower than that of benign clients. This result is consistent with the experimental results on image datasets.

\end{document}